\documentclass[11pt]{article}
\usepackage[preprint]{acl}
\usepackage{times}
\usepackage{latexsym}
\usepackage[T1]{fontenc}
\usepackage[utf8]{inputenc}
\usepackage{microtype}
\usepackage{inconsolata}
\usepackage{graphicx}
\usepackage{amsmath}
\usepackage{amssymb}
\usepackage{booktabs}
\usepackage{array}
\usepackage{enumitem}
\usepackage{float}
\usepackage{cuted} % In-place full-width main result table.
\usepackage{multirow}
\usepackage{fvextra}
\usepackage{comment}

\newcommand{\Description}[1]{}

\title{Shared Perception, Individual Decisions: A Controllable and Scalable Financial Market Simulation}

\author{
\textbf{Dejun Cao}$^{1,2}$\thanks{Equal contribution; co-first authors.}\quad
\textbf{Boyang Wang}$^{2}$\footnotemark[1]\quad
\textbf{Jiayuan Wei}$^{2}$\quad
\textbf{Guo Zhu}$^{2}$ \\
\textbf{Yishen Chen}$^{2}$\quad
\textbf{Ruixuan Yan}$^{1}$\quad
\textbf{Chengyuan Zhang}$^{1}$\quad
\textbf{Tingxi Zhang}$^{2}$ \\
\textbf{Zheng Yu}$^{2,3}$\quad
\textbf{Fan Bu}$^{2}$\quad
\textbf{Jiale Han}$^{3}$\thanks{Corresponding authors.}\quad
\textbf{Honghai Yu}$^{1}$\footnotemark[2] \\[0.4em]
$^{1}$Nanjing University \\
$^{2}$The Chinese University of Hong Kong, Shenzhen \\
$^{3}$Shenzhen Loop Area Institute \\[0.4em]
\texttt{\{boyangwang,jiayuanwei,zhuguo\}@link.cuhk.edu.cn} \\
\texttt{\{yishenchen,tingxizhang,bufan\}@link.cuhk.edu.cn} \\
\texttt{\{201098152,221870177,231098106\}@smail.nju.edu.cn} \\
\texttt{zheng.yu3@mail.mcgill.ca} \\
\texttt{jialehan@slai.edu.cn}\quad\texttt{hhyu@nju.edu.cn}
}

\begin{document}
\maketitle

\begin{abstract}
Financial market simulation models how heterogeneous investors
interact to shape market outcomes.
Recent approaches prompt large language models (LLMs) to act as
investors with different preferences, interpret market information,
and generate trading decisions.
However, combining perception and decision-making in per-agent
LLM calls makes behavioral control dependent on prompt interpretation and duplicates the processing of shared information.
We propose \textbf{SUM}, a hybrid Semantic-Utility Model that
decouples type-specific LLM perception from agent-level utility-based decisions.
LLMs generate shared signals for each investor type.
Individual agents then produce orders through explicit utility rules
using their own perception noise, preferences, portfolio states,
and constraints, preserving heterogeneity within each type.
Experiments in an A-share limit-order-book environment demonstrate SUM's effectiveness, controllability, and scalability. More broadly, our work suggests a design principle for LLM-based simulation that organizes semantic understanding and behavioral control into distinct components. Code is available at
\url{https://anonymous.4open.science/r/spid-2DE6/}.

\end{abstract}

\section{Introduction}

\begin{figure}[t]
    \centering
    \includegraphics[width=0.88\columnwidth]{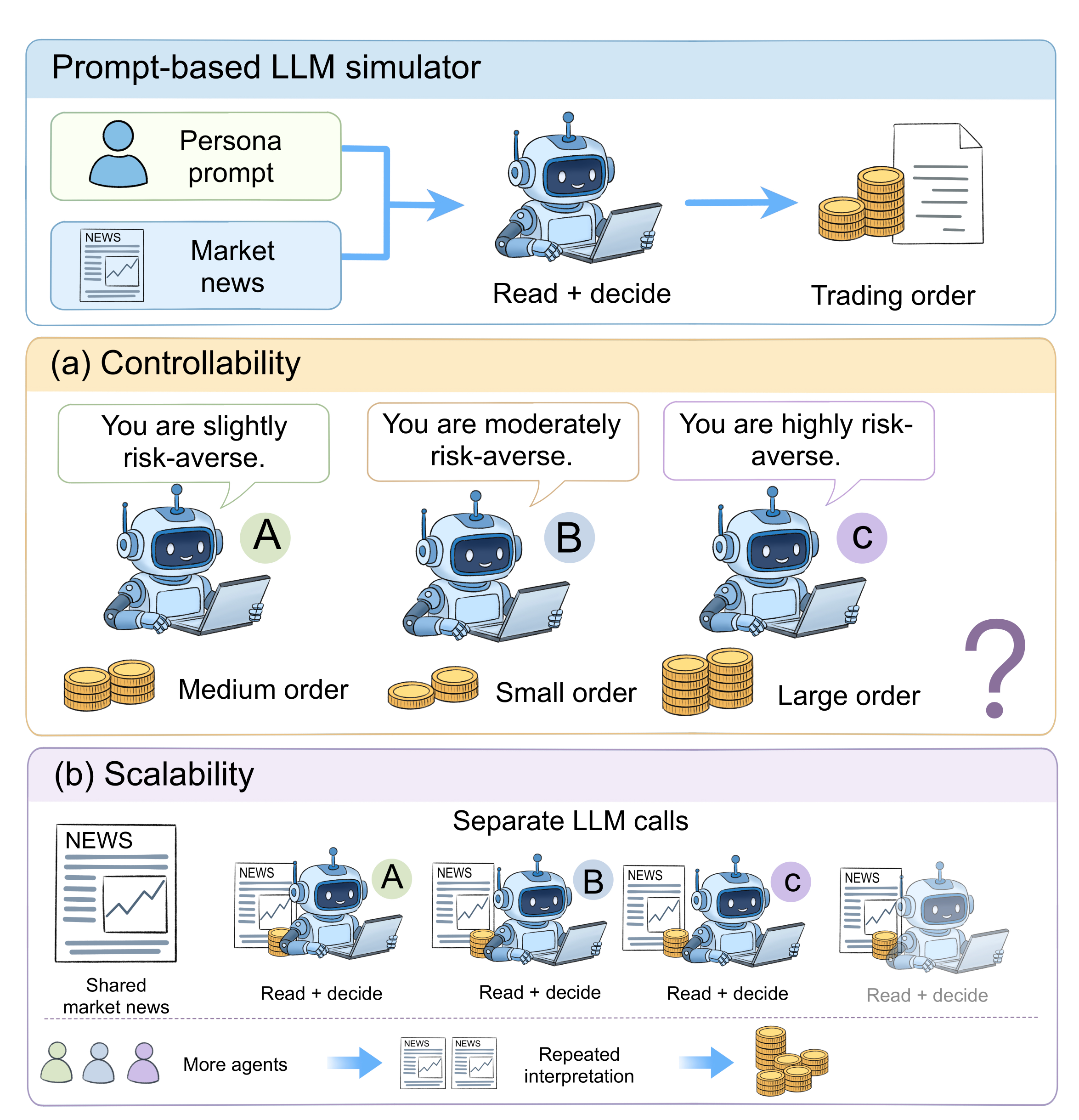}
    \caption{Challenges in LLM market simulation.}
    \label{fig:intro_challenges}
\end{figure}

Financial market simulation studies how investors with different information and preferences interact and shape market outcomes.
Traditional agent-based models represent investor behavior through explicit trading rules and utility functions~\citep{brock1998heterogeneous,chiarella2009impact}. Recent advances in large language models (LLMs) enrich these simulations
by enabling agents to interpret news, discussions, and market narratives.
Systems such as StockAgent~\citep{zhang2024stockagent} and
TwinMarket~\citep{yang2025twinmarket} use prompts to define investor
roles and guide LLMs in interpreting market information and generating
trading decisions.
This approach offers a flexible way to model how investors with
diverse perspectives respond to unstructured information and changing
market conditions.

Despite these benefits, relying on per-agent LLM inference to interpret
market information and generate trading actions raises two challenges.
First, \textit{controllability} requires a predictable relationship
between investor preferences and trading behavior.
When roles and preferences are specified through prompts, their effects
on actions depend on the LLM's interpretation, making it difficult to
ensure that a profile change produces the intended behavioral response.
Second, \textit{scalability} requires expanding investor populations
at manageable computational cost while maintaining simulation quality.
When each agent independently interprets shared market information,
population growth entails substantial repeated LLM inference.
As shown in Figure~\ref{fig:intro_challenges}, both challenges reflect the coupling of shared information
interpretation and individual action generation within per-agent
LLM calls.

To assess these challenges, we conduct a pilot study using StockAgent
and TwinMarket in a unified market environment.
With market information and agent states held fixed, increasingly
risk-averse profiles do not consistently reduce mean requested order
size.
Increasing the agent population also raises token consumption
approximately linearly, without consistent improvements in
market-tracking quality.
These findings motivate separating shared information
interpretation from individual decision-making, allowing semantic
processing to be reused while providing explicit control over
investor behavior.

To this end, we propose a hybrid Semantic-Utility Model (\textbf{SUM}) that combines cohort-level LLM perception with agent-level utility-based decisions.
The central idea is that semantic interpretation can be shared within
investor types while trading decisions remain individual.
LLMs interpret public information from each type's perspective and
produce structured signals for reuse across its members.
Each agent then combines these signals with its own perception noise,
preferences, portfolio state, and constraints to generate orders
through explicit utility rules.
This design preserves heterogeneity within investor types and places
behavioral parameters directly in the decision process.
Shared perception reduces repeated inference: with investor types,
assets, and simulation steps fixed, adding agents requires no
additional shared-perception calls.

We evaluate SUM in an A-share limit-order-book environment.
SUM improves selected return statistics and achieves lower
index-tracking error than StockAgent and TwinMarket in a six-day
comparison. Controlled interventions show consistent responses to
risk damping with lower decision variability than prompt-based
baselines. As the retail population grows from 50 to 1,000,
shared-perception calls remain fixed and token usage stays
approximately constant. Our contributions are threefold: 
\begin{itemize}
    \item We identify limitations in prompt-based behavioral control and the inference efficiency
    of LLM-based market simulators.
    \item We propose SUM, which decouples type-specific semantic perception from agent-level utility-based decisions, enabling explicit behavioral control and inference reuse while preserving individual heterogeneity.
\item Extensive experiments demonstrate the effectiveness,
controllability, and scalability of SUM in financial market simulation.
\end{itemize}

\section{Related Work and Discussion}
\label{sec:related_work}

\paragraph{Multi-Agent Social Simulation.}
%Multi-agent social simulation studies how individual behavior and interaction produce collective outcomes.
Traditional agent-based modeling uses explicit local rules to study social processes, including segregation and the formation of artificial societies~\citep{schelling1971dynamic,epstein1996growing}.
Recent LLM-based systems use language models to support communication, reasoning, and behavior generation.
Generative Agents combines memory, reflection, and planning to simulate everyday social interaction~\citep{park2023generative}.
AgentSociety supports large-scale social simulation through realistic environments and parallel execution~\citep{zhang2025parallelized}.
Within this broader field, SUM focuses on behavioral control and the cost of repeated information processing in LLM-based market simulation.

\paragraph{Agent-Based Modeling for Financial Markets.}
Agent-based finance studies how investors with different beliefs and trading rules generate market dynamics~\citep{lebaron2000agent,tesfatsion2006agent,farmer2009economy}.
Early work includes heterogeneous-agent models~\citep{brock1998heterogeneous,lux1999scaling} and the Santa Fe and Genoa artificial stock markets~\citep{arthur1997asset,raberto2001agent}.
\citet{chiarella2009impact} connect fundamentalist, chartist, and noise components to utility-based orders in a limit-order book.
Related financial theory examines noise-trader risk~\citep{delong1990noise} and information differences among investors~\citep{grossman1980impossibility,hellwig1980aggregation}, while \citet{cont2001empirical} summarize empirical market patterns used to evaluate simulations.
More recent systems, including StockAgent and TwinMarket, use LLMs to interpret market information and generate investor actions~\citep{zhang2024stockagent,yang2025twinmarket}.
Hybrid approaches also exist: \citet{hashimoto2025agent} use LLMs for trade direction and rules for order prices and volumes.
SUM builds on these modeling foundations by using shared cohort-level LLM signals as inputs to individual utility-based order generation.

\paragraph{LLM-Based Financial Decision-Making.}
LLMs are also used for financial information analysis, trading, and portfolio management.
FinGPT provides open-source financial language models~\citep{yang2023fingpt}.
Single-agent trading systems include FinMem, which uses investor profiles and layered memory, and FinAgent, which combines numerical, textual, and visual information~\citep{yu2024finmem,zhang2024finagent}.
Multi-agent frameworks organize financial decision-making through specialized roles.
FinCon uses a manager--analyst hierarchy for stock trading and portfolio management, with specialized information roles, explicit risk control, and selective belief propagation~\citep{yu2024fincon}.
These systems support investment decisions by an individual or a cooperating team, while SUM simulates market interactions among investors with separate preferences and objectives.
% Evidence: data_Oct12/02_scaling_efficiency/2025-06_top5及敏感性.
% Use the same six dates for step costs and indices, not full-month costs.
% Baselines: original N=50 week1 runs, and N=100/N=500 week1 runs.
% Institutional decisions are prerecorded separately for each N.
% Control source: 01_前置实验/gamma扫描及稳健性/data/r10_no_rb_20260526.
% TODO before submission: verify model/checkpoint and decoding settings.
% Control manifest records anon-llm; do not reuse P01's temperature here.
% Token totals include retail input and output tokens; institution cache-generation costs are excluded.
\section{Pilot Study}
\label{sec:pilot_study}

%Financial LLM systems use different designs.
% FinMem uses layered memory~\citep{yu2024finmem}, and FinAgent combines numerical, textual, and visual information~\citep{zhang2024finagent}. FinCon uses a manager--analyst hierarchy, risk control, and selective belief updates~\citep{yu2024fincon}. Market simulators include StockAgent~\citep{zhang2024stockagent}, Agent-based Simulated Financial Market (ASFM)~\citep{gao2024simulating}, and TwinMarket~\citep{yang2025twinmarket}. These systems address different tasks and should not be treated as one prompt-based design. We adapt StockAgent and TwinMarket to a common market environment and test scaling costs and agent-level risk control. The results below apply to the tested implementations and settings.

We conduct a pilot study to examine whether existing LLM-based market simulators can scale efficiently while providing reliable control over investor behavior.
We adapt StockAgent~\citep{zhang2024stockagent} and TwinMarket~\citep{yang2025twinmarket} to a unified market environment to assess the consistency of prompt-based risk control, inference costs, and market-tracking quality.

\subsection{Controllability}
\label{sec:pilot_controllability}

Prompt-based financial agents use investor profiles to guide trading decisions. We test whether increasingly risk-averse prompt profiles induce the intended decrease in order size.

\paragraph{Setting.}
We keep the market information and each agent's starting state fixed and change only the risk profile in the prompt.
We use five ordered profiles, from high risk tolerance to high risk aversion.
% Profile levels 1--5 correspond to gamma = 0.04, 0.06, 0.08, 0.10, 0.12.
% The LLM receives natural-language risk profiles, not these numeric values.
For each method, we test value, momentum, and noise investors, with ten repeated decisions per profile.
We measure absolute requested order value as a percentage of agent wealth.
The test asks a simple question: does a more cautious description consistently lead to an order that is no larger?

\begin{figure}[!htbp]
\centering
\includegraphics[width=\columnwidth]{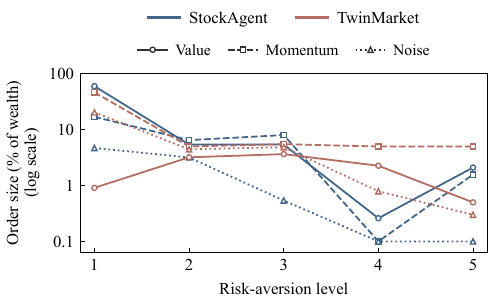}
\caption{Prompt-based risk control. Mean order size generally decreases as risk aversion increases, but five of the six method--investor combinations show reversals between adjacent profiles. Each point shows the mean absolute requested order value as a percentage of wealth over ten decisions.} % The vertical axis uses a logarithmic scale.
\label{fig:pilot_controllability}
\end{figure}

% Declare the following floats here so the two pilot figures can share a page.
\begin{figure}[!t]
\centering
\includegraphics[width=\columnwidth]{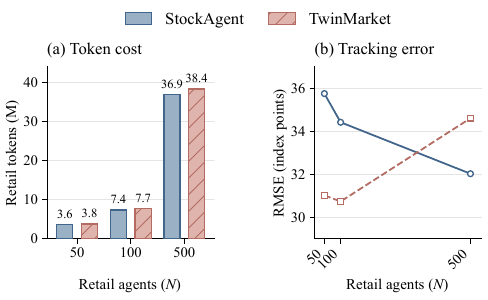}
\caption{Scaling diagnostics for StockAgent and TwinMarket over five stocks and six trading days (June 3--10, 2025). Left: input and output tokens used by retail agents; institutional cache-generation costs are excluded. Right: equal-weighted index RMSE over six daily closes, excluding initialization. } % 
\label{fig:pilot_scalability}
\end{figure}

\begin{figure*}[t]
    \centering
    \includegraphics[width=0.95\textwidth]{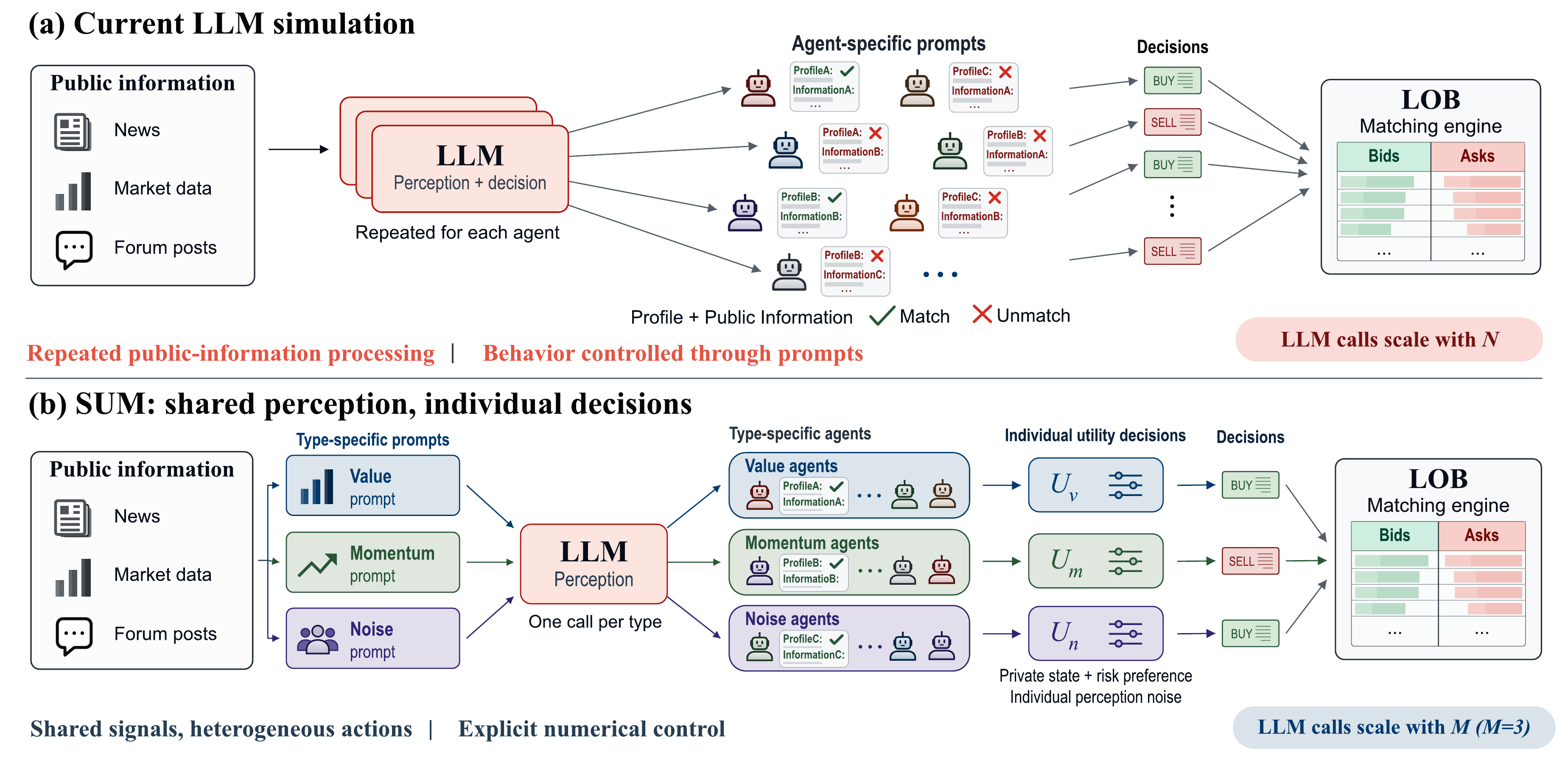}
    \captionsetup{font=footnotesize}
    \caption{The SUM pipeline. Shared public information is interpreted once
    per investor type by the LLM; individual agents then apply private
    perception noise, risk-damped utility functions, portfolio constraints,
    and LOB execution rules to generate orders.}
    \Description{Overview of the SUM pipeline: cohort-level LLM perception,
    agent-level utility decision, and LOB order submission.}
    \label{fig:framework}
\end{figure*}

\paragraph{Results.}
Five of the six method--investor combinations show at least one increase in mean order size as the profiles become more risk-averse (Figure~\ref{fig:pilot_controllability}).
For example, StockAgent's value investors increase their mean order size from 0.26\% to 2.10\% of wealth between the two most risk-averse profiles.
TwinMarket's value investors increase theirs from 0.91\% to 3.65\% between the first and third profiles.
These responses move against the intended control direction.
StockAgent's noise investors retain the expected ordering, so the problem is not that risk descriptions have no effect, but that their effect is not consistent across investor types and risk settings.

A separate test in Appendix~\ref{app:semantic_exp} shows that semantically equivalent descriptions can also produce different action distributions, with 44\% of within-level paraphrase comparisons exceeding the empirical repeat-noise threshold.

\noindent\textbf{Observation I.} \textit{Prompt-based profiles do not consistently translate specified risk preferences into predictable trading behavior, limiting fine-grained behavioral control.}

\subsection{Scalability}
\label{sec:pilot_scalability}

\paragraph{Setting.}
We compare $N\in\{50,100,500\}$ retail agents over the same five stocks and six trading days, June 3--10, 2025.
This six-day comparison is separate from the April scaling experiment and the June full-month check (Appendix~\ref{app:scaling_data_provenance}).
For each $N$, both methods replay the same prerecorded institutional decisions.
We count input and output tokens used by retail agents and measure quality by the root mean squared error (RMSE) between simulated and real equal-weighted indices, both initialized at 1,000.

\paragraph{Results.}
As shown in Figure~\ref{fig:pilot_scalability}, increasing $N$ from 50 to 500 raises token use from 3.6 to 36.9 million for StockAgent and from 3.8 to 38.4 million for TwinMarket. However, market-tracking quality does not consistently improve. Tracking error decreases with population size for StockAgent, whereas it initially decreases and then increases for TwinMarket.

\noindent\textbf{Observation II.} \textit{For the tested methods, more agents require more tokens but do not consistently improve market-tracking quality.}
%Simulation scale should therefore be assessed together with inference cost and market-tracking quality.

%These observations motivate a controllable and scalable market simulation framework that provides explicit control over individual trading behavior and reduces repeated LLM inference, enabling effective and efficient scaling to larger agent populations.

% =============================================================
%  Problem Formulation
% =============================================================
\section{Problem Formulation}
\label{sec:problemformulation}

\subsection{Market Environment and Agent Decision}
\label{sec:market_env}

We consider a market simulation in which $N$ heterogeneous agents repeatedly
interact under a shared public information structure.
At each trading step, the market state factorizes into a public component
observed by all participants and a private component restricted to each agent:
\begin{equation}
\label{eq:state}
\small
\Xi_t = \left(S_t^{\mathrm{pub}},\;\{s_{i,t}^{\mathrm{priv}}\}_{i=1}^{N}\right),
\end{equation}
where $S_t^{\mathrm{pub}}$ comprises macro news, market indices, order-book
summaries, and any other information simultaneously available to all
participants, while $s_{i,t}^{\mathrm{priv}}$ encodes agent $i$'s portfolio
holdings, capital constraints, and behavioral role attributes.

The agent decision process can be understood as two conceptually distinct
steps.
First, the agent observes and interprets public market information:
\begin{equation}
\label{eq:public_perception}
\small
\mathcal{P}\!\left(S_t^{\mathrm{pub}}\right).
\end{equation}
Second, the agent combines this public perception with private preferences and
constraints to form a trading decision:
\begin{equation}
\label{eq:private_decision}
\small
\mathcal{D}_i\!\left(\mathcal{P}\!\left(S_t^{\mathrm{pub}}\right),
s_{i,t}^{\mathrm{priv}}\right),\qquad
\mathcal{D}_i\!\left(\cdot\right) \in \mathcal{A}_i .
\end{equation}
Here $\mathcal{P}$ denotes perception over shared public information, while
$\mathcal{D}_i$ denotes the private decision rule induced by agent-specific
preferences, holdings, capital constraints, and role attributes.
This decomposition separates what is common to many agents from what must
remain individualized.

\subsection{Existing Per-Agent LLM Decision Architecture}
\label{sec:agent_decision}

Existing LLM-based market simulators usually collapse the two steps above into
one single-agent, single-step LLM decision.
For every agent at every trading step, the LLM jointly reads the public
information and the private profile, then directly emits the trading decision:
\begin{equation}
\label{eq:fused_agent_decision}
\small
\mathrm{LLM}_{i,t}^{\mathrm{decision}}
=
\mathrm{LLM}\!\left(
\mathcal{P}\!\left(S_t^{\mathrm{pub}}\right)
+
\mathcal{D}_i\!\left(s_{i,t}^{\mathrm{priv}}\right)
\right).
\end{equation}

In this architecture, public semantic perception and private preference-based
decision-making are not modularized.
They are fused inside the same language-model inference, so the public
information must be re-read and re-interpreted independently for every agent.
The resulting per-step LLM complexity is:

\begin{equation}
\label{eq:fused_llm_complexity}
\small
\begin{aligned}
C_{\mathrm{LLM}}
&= N\cdot \mathrm{LLM}(\mathcal{P}+\mathcal{D}) \\
&= N\cdot \mathrm{LLM}(\mathcal{P})
+ N\cdot \mathrm{LLM}(\mathcal{D}) .
\end{aligned}
\end{equation}
The first term is redundant because $S_t^{\mathrm{pub}}$ is shared across the population, yet the architecture pays for public perception $N$ times.
The second term is unavoidable only to the extent that agents have different
private states and preferences.
This fused design therefore makes the cost of public-information processing
grow linearly with the number of agents. %, which is the core scalability bottleneck addressed by SUM.

% =============================================================
%  Methodology: SUM Architecture
% =============================================================
\section{The SUM Framework}
\label{sec:methodology}

\subsection{Overall Framework: Decoupling Perception and Decision}
\label{sec:sum_architecture}

The per-agent formulation in Section~\ref{sec:agent_decision} combines
public information interpretation and private decision-making within
each agent's LLM call.
As shown in Figure~\ref{fig:framework}, SUM separates these operations into type-specific semantic perception and agent-level utility-based decisions.
In the retail-agent pipeline, investors of the same type share an
LLM-generated interpretation of public market information, while each
investor generates orders using its own perception noise, preferences,
and portfolio state.

Let $N$ denote the number of retail agents, partitioned into $M$
investor-type cohorts $\mathcal{M}=\{1,\ldots,M\}$, with $c(i)$ denoting
the type of agent $i$.
At each perception update, SUM invokes the LLM once per cohort and
reuses the resulting signal across its members.
Individual decisions are then computed through explicit utility rules.
For a fixed set of assets, the computational costs of these two
components are
\begin{equation}
\label{eq:cost_sum}
\small
\begin{aligned}
C^{\mathrm{SUM}}= M C_{\mathcal{P}} + N C_{\mathcal{U}}, \quad
C_{\mathrm{LLM}}^{\mathrm{SUM}}= M C_{\mathcal{P}},
\end{aligned}
\end{equation}
where $C_{\mathcal{P}}$ is the average cost of a cohort-level LLM
perception call and $C_{\mathcal{U}}$ is the average cost of an
agent-level utility-based decision.
Compared with Eq.~\ref{eq:fused_llm_complexity}, the number of LLM
calls depends on the number of investor types rather than the number
of agents.
With fixed investor types and market inputs, increasing $N$ therefore
requires no additional perception calls, although individual decision
computation and market execution still grow with population size.

This separation connects scalability and controllability:
shared perception reduces repeated inference, while explicit utility
rules allow individual preferences to directly govern trading behavior.
Agent-specific signals, parameters, and private states preserve
heterogeneity within each cohort.

\subsection{Type-Specific LLM-Based Perception}
\label{sec:investor_design}

% The main comparison table is placed in the Experiments section.

The perception layer interprets public market information from the
perspective of each investor type.
For cohort $m$ at time $t$, the shared signal is
\begin{equation}
\label{eq:sum_perception}
\small
\bar{P}_{m,t}
=
\mathcal{P}_{\Theta}\!\left(S_t^{\mathrm{pub}};\,q_m\right),
\end{equation}
where $S_t^{\mathrm{pub}}$ denotes public market information,
$\mathcal{P}_{\Theta}$ is the LLM-based perception operator, and
$q_m$ is a type-specific prompt.
The output $\bar{P}_{m,t}$ contains structured semantic scores for
subsequent utility evaluation.
SUM uses three retail investor types, each emphasizing a different
set of market relationships. The complete perception prompts and output schemas are provided in Appendix~\ref{app:sum_retail_prompts}.

\noindent\textbf{Value investors.}
Value perception assesses the relationship between market prices and
fundamentals.
Its signals summarize valuation levels, earnings sustainability,
asset quality, macroeconomic risk, and relevant evidence from news
and forum discussions.

\noindent\textbf{Momentum investors.}
Momentum perception assesses recent price-volume movements and
information diffusion.
Its signals summarize trend strength, moving-average direction,
range position, volume confirmation, and event catalysts.
They also distinguish catalysts that are still diffusing from those
already attracting crowded attention.

\noindent\textbf{Noise investors.}
Noise perception captures short-horizon stimuli and sentiment pressure.
Its signals summarize forum sentiment, immediate reactions to news,
order-book imbalance, volume spikes, and recent price shocks.

\subsection{Agent-Level Utility-Based Decisions}
\label{sec:heterogeneity}

The decision layer combines shared cohort signals with agent-specific
perception noise, preferences, and portfolio states, enabling investors
of the same type to make distinct trading decisions.

\paragraph{Agent-specific signal construction.}
Each agent first forms an individual interpretation of its cohort's
shared signal:
\begin{equation}
\label{eq:sum_private_signal}
\small
P_{i,t}
=
\bar{P}_{c(i),t}+\epsilon_{i,t},\qquad
\epsilon_{i,t}\sim\mathcal{N}\!\left(0,\sigma_i^2 I\right).
\end{equation}
The parameter $\sigma_i$ controls the magnitude of perception noise,
with $\sigma_i\sim\mathcal{U}[0.005,0.030]$ in our implementation.
Smaller values keep the agent's interpretation closer to the shared
signal, while larger values introduce greater individual variation.
This step preserves differences in perception without additional
LLM calls.

\paragraph{Utility-based order generation.}
The personalized signal is combined with the agent's private state
and preferences to determine its trading action:
\begin{equation}
\label{eq:sum_decision}\small
\begin{aligned}
z^*_{i,t}
&=\arg\max_{|z|<w_{c(i)}}
\mathcal{U}_{c(i)}
\!\left(z;\,P_{i,t},s_{i,t}^{\mathrm{priv}},\gamma_i\right),\\
a_{i,t}
&=\operatorname{Order}_{\mathcal{A}_i}
\!\left(z^*_{i,t};s_{i,t}^{\mathrm{priv}}\right).
\end{aligned}
\end{equation}
where $\mathcal{U}_{c(i)}$ is the type-specific utility function,
$s_{i,t}^{\mathrm{priv}}$ contains cash, holdings, and recent
portfolio history, and $\gamma_i$ is the risk-damping parameter.
The feasible action set $\mathcal{A}_i$ incorporates the agent's
portfolio and trading constraints.

The utility rules translate type-specific semantic signals into
expected-return scores that guide order generation.
For value investors, the rules favor undervaluation and positive
fundamental revisions while penalizing weak asset quality and
unfavorable macroeconomic information.
For momentum investors, they favor trends supported by price-volume
evidence and catalysts that are still diffusing.
For noise investors, positive and negative sentiment pressure support
buying and selling, respectively, with signal magnitude determining
trading intensity before risk damping and feasibility constraints.

\paragraph{Preference parameters and behavioral control.}
Individual risk preferences are specified through
$\gamma_i\sim\mathcal{U}[0.04,0.12]$.
Holding perception signals and private states fixed, larger
$\gamma_i$ dampens the intended position adjustment.
The parameter therefore provides an explicit means of controlling
how strongly an agent trades on a given signal.
Additional type-specific parameters, including $\psi_i$, govern
mean-reversion beliefs, trend-following weights, and sentiment
persistence.

These controls separate variation in interpretation from variation
in trading intensity: $\sigma_i$ determines how an agent perceives
the shared signal, while $\gamma_i$ determines how strongly it
responds.
Together with differences in cash and holdings, they allow investors
within the same cohort to generate distinct orders.

\paragraph{Order execution and state updates.}
Feasible orders are submitted to the limit-order book.
Execution updates each agent's cash and holdings, while the resulting
market prices inform subsequent perception and decision steps.

\begin{comment}
SUM does not create scale by collapsing each investor family into one
representative agent.
Instead, agents of the same type share the expensive perception result but
remain heterogeneous through two continuous controls.

\paragraph{Ability heterogeneity.}
The parameter $\sigma_i$ governs perception noise in
Eq.~\ref{eq:sum_private_signal}.
In our implementation, $\sigma_i\sim\mathcal{U}[0.005,0.030]$.
Lower $\sigma_i$ means the agent tracks the cohort semantic signal more
precisely; higher $\sigma_i$ introduces stronger idiosyncratic interpretation
noise.

\paragraph{Preference heterogeneity.}
The parameter $\gamma_i$ enters the utility layer in
Eq.~\ref{eq:sum_decision} as a risk-damping coefficient, with
$\gamma_i\sim\mathcal{U}[0.04,0.12]$.
Larger $\gamma_i$ reduces aggressive position adjustment under the same
expected-return signal.
Investor-type persistence is controlled by $\psi_i$: value agents use a
higher mean-reversion belief, momentum agents use a higher trend-following
weight, and noise agents use a lower-persistence sentiment process.
Thus, $\sigma_i$ controls \textit{how agents perceive} shared information,
while $\gamma_i$ controls \textit{how strongly they trade} on the resulting
signal.
\end{comment}

% The framework figure follows the balanced Pilot Study ending in pilot_study.tex.

Full retail perception prompts and output schemas are provided
in Appendix~\ref{app:sum_retail_prompts}.
The corresponding utility functions and heterogeneity parameters
are detailed in Appendix~\ref{app:value},
Appendix~\ref{app:momentum}, and Appendix~\ref{app:noise}.

% Experiments: simulation quality, controllability, and scalability.
\section{Experiments}
\label{sec:experiments}

% Main comparison: separate panels retain their original experimental scopes.
\begin{table*}[t]
  \centering
  \fontsize{10pt}{12pt}\selectfont
  \renewcommand{\arraystretch}{0.85}
  \setlength{\aboverulesep}{0.2ex}
  \setlength{\belowrulesep}{0.25ex}
  \scalebox{0.90}{%
  \begin{minipage}{\textwidth}
  \centering
  \begin{tabular*}{0.98\textwidth}{@{\extracolsep{\fill}}llrrrr@{}}
    \toprule
    \multicolumn{6}{@{}l}{\textbf{(a) Selected market-level results: H1 2025, 50 stocks, 117 trading days}} \\
    \midrule
    \textbf{Metric} & \textbf{SUM run / $N$} & \textbf{SUM}
      & \shortstack{\textbf{StockAgent}\\[-1pt]$N=500$}
      & \shortstack{\textbf{TwinMarket}\\[-1pt]$N=500$}
      & \shortstack{\textbf{Rule-based}\\[-1pt]$N=1000$} \\
    \midrule
    RetCorr $\uparrow$ & H1 slice / 1000 & $\mathbf{0.264}$ & $0.250$ & $0.201$ & $-0.089$ \\
    SD error (pp) $\downarrow$ & H1 slice / 1000 & $\mathbf{0.718}$ & $0.952$ & $0.970$ & $0.902$ \\
    Return W1 (pp) $\downarrow$ & Cached replay / 100 & $\mathbf{0.526}$ & $0.573$ & $0.594$ & $0.534$ \\
    \bottomrule
  \end{tabular*}
  \par\vspace{1pt}
  \begin{tabular*}{0.94\textwidth}{@{\extracolsep{\fill}}lrrrr@{}}
    \toprule
    \multicolumn{5}{@{}l}{\textbf{(b) Index-tracking RMSE\,$\downarrow$: June 3--10, 2025, five stocks, six trading days}} \\
    \midrule
    \textbf{System} & \textbf{$N=50$} & \textbf{$N=100$} & \textbf{$N=500$} & \textbf{$N=1000$}$^{\dagger}$ \\
    \midrule
    StockAgent & 35.8 & 34.4 & 32.0 & --- \\
    TwinMarket & 31.0 & 30.7 & 34.6 & --- \\
    \textbf{SUM} & \textbf{24.9} & \textbf{27.2} & \textbf{22.2} & 33.0 \\
    \bottomrule
  \end{tabular*}
  \end{minipage}%
  }
  \captionsetup{font=10pt,justification=raggedright,singlelinecheck=false}
  \caption{Simulation quality under two protocols.
    \textbf{(a)} Selected H1 metrics against the StockAgent run's real
    equal-weight index. RetCorr: daily return correlation;
    SD error: absolute daily standard-deviation error;
    W1: return-distribution Wasserstein distance; pp: percentage points.
    SUM rows use different runs and agent counts
    (Appendix~\ref{app:h1_market_comparison}).
    \textbf{(b)} Equal-weighted index RMSE over six daily closes, excluding
    initialization; one run per setting, seed 42.
    Bold: best reported values per row in (a) and at matched agent counts
    in (b); arrows: preferred directions.}
    %$^{\dagger}$SUM only; dashes: not evaluated.
  \label{tab:main_market_results}
  \label{tab:ew_quality_scale}
\end{table*}

In this section, we answer three questions:
1) How well does SUM simulate financial markets?
We evaluate selected return metrics over the first half of 2025
(H1) against real observations and compare index-tracking errors
in a separate six-day experiment (Section~\ref{sec:fidelity}).
2) Can SUM provide reliable control over agent behavior?
We test whether changes in risk preferences produce ordered actions and
whether actions remain consistent across repeated decisions
(Section~\ref{sec:controllability}).
3) How does SUM scale as the number of agents increases?
We compare LLM calls and token use at different agent counts
(Section~\ref{sec:scalability}).

\paragraph{Baselines and model.}
We adapt StockAgent~\citep{zhang2024stockagent} and
TwinMarket~\citep{yang2025twinmarket} to our market setting and include a
rule-based baseline in the H1 comparison.
We use Qwen3-8B with temperature $=0.6$, \texttt{top\_p}$=0.95$, and
\texttt{top\_k}$=20$.
Baseline implementations and detailed experimental settings are provided in
Appendices~\ref{app:baselines} and~\ref{app:main_experiment_setups}.

% E1: Simulation quality.
\subsection{Simulation Quality}
\label{sec:fidelity}

\paragraph{Setup and metrics.}
Table~\ref{tab:main_market_results}(a) compares archived runs over
January--June 2025 (50 MSCI China A50 stocks; 117 trading days).
StockAgent and TwinMarket use 500 retail agents, and Rule-based uses 1,000.
SUM uses the H1 slice of a 1,000-agent annual run for return correlation
and daily standard-deviation error, and a separate 100-agent cached replay
for return-distribution distance (W1). All runs use a common real-index
reference. These metrics and SUM runs were selected after examining the
archived results. Appendix~\ref{app:h1_market_comparison} gives the
definitions and settings; the replay is not used to estimate inference cost.

\paragraph{Market-level comparison.}
The 1,000-agent SUM run has a return correlation of 0.264, compared with
0.250 for StockAgent, 0.201 for TwinMarket, and $-0.089$ for Rule-based.
Its reported daily standard-deviation error is also the lowest, at
0.718 percentage points. The separate 100-agent SUM replay has the lowest
reported W1, at 0.526 percentage points. These results show advantages on
the selected measures of return co-movement and distribution. They do not
establish that one SUM configuration is better on all fidelity measures,
and the H1 comparison does not hold agent counts fixed.

\paragraph{Index tracking at matched agent counts.}
We also compare the three LLM-based systems at the same agent counts in a
separate five-stock experiment covering June 3--10, 2025.
Table~\ref{tab:main_market_results}(b) reports equal-weighted index RMSE over
six daily closes. SUM has the lowest RMSE at all three shared scales,
$N=50,100,500$. At $N=500$, its RMSE is 22.2, compared with 32.0 for
StockAgent and 34.6 for TwinMarket, a reduction of 30.6\% and 35.8\%,
respectively. This supports better index tracking in this setting, but
each setting uses one run with seed 42.

\begin{figure*}[t]
  \centering
  \includegraphics[width=0.88\textwidth]{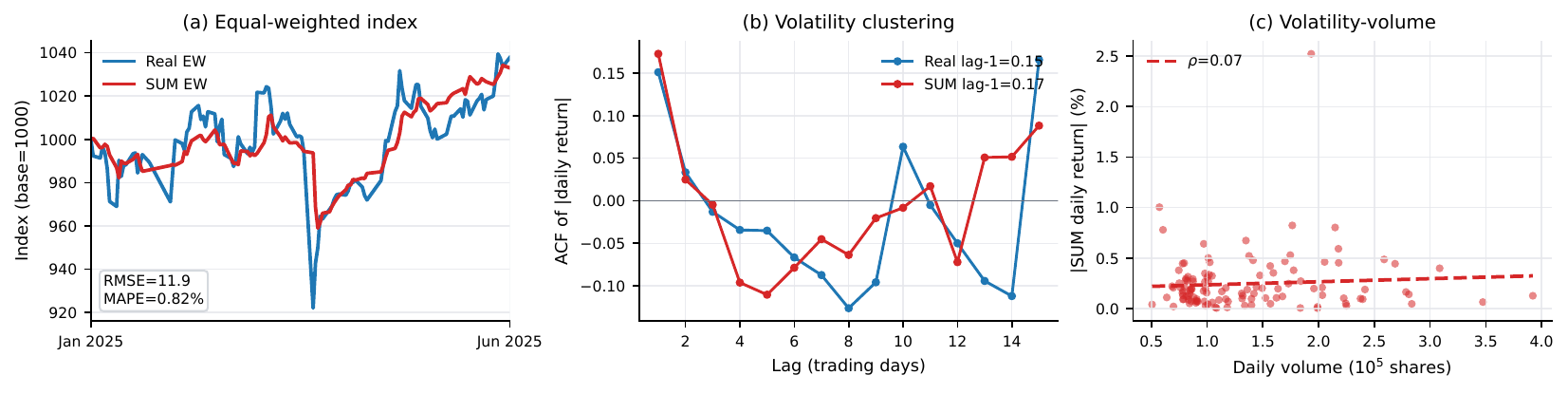}
  \captionsetup{font=footnotesize}
  \caption{H1 2025 diagnostics (January--June; 117 trading days).
  \textbf{(a)}~SUM and real equal-weighted index paths, normalized to 1,000.
  \textbf{(b)}~Absolute-return autocorrelation (lag 1: SUM 0.173; real 0.151).
  \textbf{(c)}~SUM daily volume versus absolute return (Pearson $r=0.070$).
  This archived diagnostic is a different run from the H1 comparison in
  Table~\ref{tab:main_market_results}(a), despite the shared calendar window;
  no rule-based baseline is plotted.}
  \Description{Three panels for the H1 2025 diagnostic: SUM and real
  equal-weighted index paths, absolute-return autocorrelations, and SUM
  volume versus absolute daily return.}
  \label{fig:emergent_fidelity}
\end{figure*}

\paragraph{Additional diagnostics.}
Figure~\ref{fig:emergent_fidelity} reports a separate archived H1 2025
diagnostic, not the SUM runs in Table~\ref{tab:main_market_results}(a).
SUM and the real market have similar lag-1 absolute-return autocorrelations
(0.173 and 0.151), but daily return correlation is $-0.130$.
Full-year investor records also show different returns across investor
types; detailed results are provided in Appendix~\ref{sec:fidelity_appendix}.
These diagnostics describe selected market and investor patterns, not an
overall advantage in market realism. Data availability is described in
Appendix~\ref{app:fidelity_data_provenance}; separate heterogeneity
interventions are reported in Appendix~\ref{sec:ablation}.

% E2: Controllability.
\subsection{Controllability}
\label{sec:controllability}

\paragraph{Setup.}
We extend the fixed-state pilot in Section~\ref{sec:pilot_controllability}
by adding SUM to the StockAgent and TwinMarket comparison.
The single-asset task uses three investor types and five risk levels. Each
method--investor--level combination has ten repeated decisions.
SUM uses a numerical risk-damping parameter, while the baselines express
the corresponding risk preference through prompt profiles.
More details are in Appendix~\ref{app:ctrl_setup}.

\paragraph{Metrics.}
We measure absolute requested order value relative to initial wealth,
$\Delta_i(\gamma)=|\mathrm{deal\_amount}_i|/W_0$.
Spearman correlation measures whether mean order size decreases as risk
aversion increases. Adjacent inversion rate (AIR) measures violations of
this ordering between neighboring levels. Mean within-level sample
standard deviation measures variation across repeated decisions.

\begin{table}[!htbp]
  \centering
  \footnotesize
  \setlength{\tabcolsep}{4pt}
  \renewcommand{\arraystretch}{1.08}
  \begin{tabular}{llccc}
    \toprule
    \textbf{System} & \textbf{Investor}
      & $\boldsymbol{\rho_s}$
      & \textbf{AIR}
      & $\bar{\boldsymbol{\sigma}}$ \\
    \midrule
    \multirow{3}{*}{StockAgent}
      & Value    & $-0.90$ & $0.25$ & $0.013$ \\
      & Momentum & $-0.80$ & $0.50$ & $0.060$ \\
      & Noise    & $-0.97$ & $0.00$ & $0.008$ \\
    \midrule
    \multirow{3}{*}{TwinMarket}
      & Value    & $-0.30^{\dagger}$ & $0.50$ & $0.017$ \\
      & Momentum & $-0.67$ & $0.25$ & $0.058$ \\
      & Noise    & $-0.90$ & $0.25$ & $0.045$ \\
    \midrule
    \multirow{3}{*}{\textbf{SUM}}
      & Value    & $\mathbf{-1.00}$ & $\mathbf{0.00}$ & $\mathbf{0.001}$ \\
      & Momentum & $\mathbf{-1.00}$ & $\mathbf{0.00}$ & $\mathbf{0.011}$ \\
      & Noise    & $\mathbf{-1.00}$ & $\mathbf{0.00}$ & $\mathbf{0.001}$ \\
    \bottomrule
  \end{tabular}
  \captionsetup{font=10pt}
  \caption{Controllability (five $\gamma$ levels; ten repeats each).
  $\rho_s$: Spearman correlation of $\gamma$ with mean $\Delta$
  (target: negative); AIR: adjacent inversion rate (target: 0);
  $\bar{\sigma}$: mean within-level sample SD (target: low).
  $^{\dagger}p=0.68$ (two-sided exact permutation test; 120 permutations).}
  \label{tab:ctrl_metrics}
\end{table}

\paragraph{Results.}
Table~\ref{tab:ctrl_metrics} shows that SUM produces decreasing mean order
sizes across all five levels for all three investor types, with
$\rho_s=-1.00$ and AIR$=0$.
StockAgent and TwinMarket generally follow the expected direction, but
five of the six method--investor combinations contain local reversals.
Their within-level standard deviations are up to 50.4 times those of SUM
for the same investor type, using unrounded values.
Thus, explicit utility parameters provide more ordered and consistent
responses in this single-step test than prompt-based risk profiles.
This tests the implemented control mechanism, not whether its preferences
fully describe real investors.

% E3: Scalability.
\subsection{Scalability}
\label{sec:scalability}

\begin{figure}[tbp]
  \centering
  \includegraphics[width=\columnwidth]{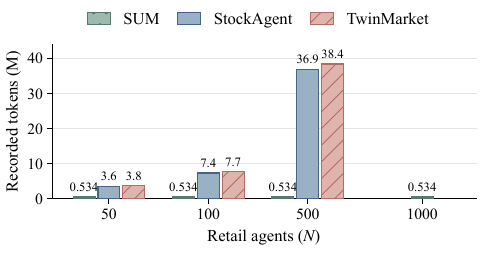}
  \captionsetup{font=10pt}
  \caption{Recorded token cost over six trading days (June 3--10, 2025;
  five stocks; one run per setting). SUM includes cohort and
  institutional inference; baseline counts cover retail inference and
  exclude institutional cache generation. No baseline run is available at
  $N=1000$.}
  \label{fig:resource_cost}
\end{figure}

\paragraph{Setup.}
We measure inference costs in the same six-day, five-stock experiment used
for the matched-agent comparison in Section~\ref{sec:fidelity}.
All three systems are evaluated at $N=50,100,500$, with an additional SUM
run at $N=1000$. Detailed settings and accounting scopes are provided in
Appendix~\ref{app:scaling_data_provenance}.

\paragraph{Results.}
Figure~\ref{fig:resource_cost} shows that SUM's recorded token use remains
near 0.534 million as $N$ increases from 50 to 1,000, while StockAgent and
TwinMarket grow approximately linearly over the measured range.
At $N=500$, their recorded token totals are 69.0 and 71.9 times SUM's,
respectively. These totals have different scopes: SUM includes cohort and
institutional inference, while baseline totals include retail inference
but exclude institutional cache generation.

SUM requires 90 cohort-perception calls at every evaluated scale.
At $N=500$, each baseline records 10,710 retail decision calls, or 119 times
this cohort-perception count. %This is not a comparison of all LLM calls.

\paragraph{Quality and scale.}
Lower inference growth does not mean that adding agents always improves
simulation quality. Table~\ref{tab:main_market_results}(b) shows that SUM's
RMSE increases from 22.2 at $N=500$ to 33.0 at $N=1000$.
The results therefore support efficient reuse of shared perception, while
the effect of agent count on simulation quality remains dependent on the
setting. The quality-per-token metric is defined in
Appendix~\ref{app:resource_efficiency}.

\section{Conclusion}
\label{sec:conclusion}

SUM combines shared LLM perception with individual utility-based decisions for controllable and scalable market simulation.
In our experiments, SUM shows more consistent risk responses and less repeated inference.
It also achieves lower index-tracking error than StockAgent and TwinMarket at matched agent counts.
These results support separating semantic perception from behavioral control.
Future work should test broader markets and compare LLM perception with rule-based and statistical alternatives.

\section*{Limitations}

Our evaluation covers a limited range of population sizes and market conditions. Moreover, investor types and utility functions are manually specified, and the demonstrated control over selected behavioral dimensions does not establish the realism of individual investor behavior. Moreover, SUM has not been validated for live trading, investment advice, or market-impact prediction. Further experiments are needed to assess SUM at larger scales and establish its robustness across different market settings.

\bibliography{references}

\clearpage
\twocolumn
\appendix
% Consolidated appendix source. Loaded once by main.tex after \appendix.
% Reorganized on 2026-11-25: A = original P/K/L/M/B.2.1; B = A.1/B.1/N/E/F/I/G; C = A.2/A.3.
% New A--C titles are intentionally blank for the authors to fill in.
% Original files are kept in archive/appendix_sources.

% ==============================================================================
% Part 1: originally appendix.tex
% ==============================================================================
% Restructured appendix.
% Required additions to the document preamble:
%   \usepackage{fvextra}
% The existing amsmath package is used for mathematical notation.
% Insert after the existing \appendix command; do not add a second \appendix.
% Standard numbering is A.1.1, A.1.2, ..., A.2.1, ..., A.3.

\setcounter{secnumdepth}{3}

% Appendix A: original P, K, L, M, and B.2.1.
\section{} % TODO(author): fill in title.

% Original P: Resource-Normalised Efficiency
\subsection{} % TODO(author): fill in title.
\label{app:resource_efficiency}

To jointly assess cost and quality we define the \emph{quality-per-token}
efficiency $\eta$ as

\begin{equation}
  \eta = \frac{1/\text{EW-RMSE}}{\text{Tok.~(M)}},
  \label{eq:efficiency}
\end{equation}

so that a system with both lower RMSE and lower token cost achieves a higher
$\eta$.
Table~\ref{tab:efficiency} reports $\eta$ for all executed runs.

\begin{table}[H]
  \centering
  \scriptsize
  \setlength{\tabcolsep}{3.5pt}
  \renewcommand{\arraystretch}{1.04}
  \begin{tabular}{@{}llrrr@{}}
    \toprule
    \textbf{System} & \textbf{$N$}
      & \textbf{EW-RMSE} & \textbf{Tok.~(M)}
      & \textbf{$\eta\times 10^{3}$}$\uparrow$ \\
    \midrule
    SUM & 50   & \textbf{24.9} & \textbf{0.534} & \textbf{75.2} \\
    SUM & 100  & \textbf{27.2} & \textbf{0.534} & \textbf{68.8} \\
    SUM & 500  & \textbf{22.2} & \textbf{0.534} & \textbf{84.2} \\
    SUM & 1000 &         33.0  & 0.534 &         56.7  \\
    \midrule
    SA  & 50   & 35.8 &  3.585 & 7.8 \\
    SA  & 100  & 34.4 &  7.355 & 3.9 \\
    SA  & 500  & 32.0 & 36.858 & 0.8 \\
    \midrule
    TM  & 50   & 31.0 &  3.756 & 8.6 \\
    TM  & 100  & 30.7 &  7.670 & 4.2 \\
    TM  & 500  & 34.6 & 38.365 & 0.8 \\
    \bottomrule
  \end{tabular}
  \captionsetup{font=footnotesize, skip=4pt}
  \caption{Quality per recorded token over June 3--10, 2025 (higher is better).
  SUM includes cohort and institutional inference; baselines exclude institutional cache generation.
  At $N=500$, SUM's efficiency is 99.4 times StockAgent's and 111.8 times TwinMarket's.
  Ratios use unrounded values.}
  \label{tab:efficiency}
\end{table}

SUM has the highest quality per recorded token at each shared scale, and its relative advantage increases over the measured population sizes.
SUM's own efficiency is not monotone in $N$: it falls from 50 to 100 agents, peaks at 500, and falls at 1000.
These single-seed, six-day results do not establish a general optimal population size or identify the cause of these changes.

% =============================================================
% Original K: Value Investor Structure
\subsection{} % TODO(author): fill in title.
\label{app:value}

The value investor archetype is grounded in the mean-reversion tradition of asset pricing. Its utility function builds on the book-to-market factor \citep{fama1992cross, fama1993common}, the AR(1) valuation framework of \citet{fama1988dividend} and \citet{campbell1988stock}, the residual income dynamics of \citet{ohlson1995earnings}, and the financial statement quality signals of \citet{piotroski2000value}. Loss-averse risk adaptation follows \citet{barberis2001mental}, and ability noise follows \citet{daniel1998investor}.

The quarterly and daily perception prompts for value investors,
together with their output schemas, are provided in
Appendix~\ref{app:sum_retail_prompts}.

\subsubsection{Utility Function}

The valuation state is computed daily without the LLM:
\[
v_t = \ln\!\left(\frac{B}{P_t}\right) = \ln\!\left(\frac{\texttt{bps}}{P_t}\right).
\]

The fundamental revision term aggregates LLM signals:
\[
\begin{aligned}
\delta_{\mathrm{revision}}
&=\alpha_{\mathrm{earn}}\cdot S_{\mathrm{earn}}
 +\alpha_{\mathrm{qual}}\cdot S_{\mathrm{qual}}\\
&\quad+\alpha_{\mathrm{macro}}\cdot S_{\mathrm{macro}},
\end{aligned}
\]
where \(\alpha_{\mathrm{earn}}=0.40\), \(\alpha_{\mathrm{qual}}=0.30\), \(\alpha_{\mathrm{macro}}=0.30\), and \(S_{\mathrm{earn}} = \mathrm{clip}(S_{\mathrm{earn}}^{\mathrm{base}} + \Delta_{\mathrm{earn}},\, -1,\, 1)\), with \(S_{\mathrm{qual}}\) defined analogously.

The expected return follows an AR(1) mean-reversion framework \citep{fama1988dividend, campbell1988stock}:
\[
\hat{R}_{t+1} = (1-\psi_i)(v_t - \bar{v}) + \delta_{\mathrm{revision}} + g_B,
\]
where \(\psi_i\) is the agent-specific mean-reversion belief and \(g_B = 0.7\, r_{10Y} + 0.3\, r_{\mathrm{LPR}}\) is the risk-free benchmark rate.

\[
\begin{gathered}
\mu_{i,t}=\hat R_{t+1}+\epsilon_{i,t},\qquad
\epsilon_{i,t}\sim\mathcal N(0,\sigma_i^2),\\
\tilde\gamma_i=\max(\gamma_i,0.001),\\
b_{i,t}=\tanh\!\left(|\mu_{i,t}|/\tilde\gamma_i\right),\\
H(u)=u\operatorname{artanh}(u)
 +\tfrac12\ln(1-u^2),\quad |u|<1.
\end{gathered}
\]
\[
\begin{aligned}
w_{\mathrm{value}}&=0.60,\\
\mathcal U_{\mathrm{value}}(z;\cdot)
 &=\mu_{i,t}z-\tilde\gamma_i w_{\mathrm{value}}
 H\!\left(\frac{z}{w_{\mathrm{value}}}\right),\\
z^*_{i,t}&=w_{\mathrm{value}}
 \tanh\!\left(\frac{\mu_{i,t}}{\tilde\gamma_i}\right),\\
\delta^{\mathrm{value}}_{i,t}
 &=\operatorname{sign}(\mu_{i,t})
 (-0.010+0.005b_{i,t}).
\end{aligned}
\]
\[
\begin{aligned}
W_{i,t}&=c_{i,t}+\sum_s h_{i,s,t}p_{s,t},\\
D^{\mathrm{raw}}_{i,t}&=W_{i,t}|z^*_{i,t}|,\\
d_{i,t}&=\begin{cases}
\mathrm{BUY},&\mu_{i,t}>0,\\
\mathrm{SELL},&\mu_{i,t}\le0,
\end{cases}\\
D_{i,t}&=\begin{cases}
\min(D^{\mathrm{raw}}_{i,t},0.95c_{i,t}),&d_{i,t}=\mathrm{BUY},\\
\min(D^{\mathrm{raw}}_{i,t},h_{i,s,t}p_{s,t}),&d_{i,t}=\mathrm{SELL}.
\end{cases}
\end{aligned}
\]
\[
\begin{aligned}
&p^{\mathrm{lim}}_{i,t}=p^{\mathrm{prev\ close}}_{s,t}
 (1+\delta^{c(i)}_{i,t}),\\
&o_{i,t}=(d_{i,t},p^{\mathrm{lim}}_{i,t},D_{i,t}),\\
&\operatorname{Order}_{\mathcal A_i}(z^*_{i,t};s_{i,t}^{\mathrm{priv}})\\
&\quad=\begin{cases}
\mathrm{HOLD},&p_{s,t}\le0\ \text{or}\ D_{i,t}<5p_{s,t},\\
\operatorname{Filter}_{\mathcal A_i}(o_{i,t}),
 &\text{otherwise},
\end{cases}\\
&c(i)\in\{\mathrm{value},\mathrm{momentum},\mathrm{noise}\}.
\end{aligned}
\]

\subsubsection{Heterogeneity Parameters}

Preference heterogeneity: \(\psi_i \sim \mathrm{Beta}(5,\, 2)\), \(\psi_i \in [0.50,\, 0.95]\). \\
Ability heterogeneity: \(\sigma_i \sim \mathrm{Uniform}(0.005,\, 0.030)\), independent of \(\psi_i\).

% ============================================================
% Original L: Momentum Investor Structure
\subsection{} % TODO(author): fill in title.
\label{app:momentum}

The momentum investor archetype is grounded in the trend-continuation literature. Its design draws on the price momentum anomaly \citep{jegadeesh1993returns}, the intermediate-horizon momentum premium of \citet{novy2012momentum}, the gradual information diffusion model of \citet{hong1999unified}, volume--return interactions \citep{lo2000trading}, and the 52-week high anchor of \citet{george2004high}.

The daily catalyst and diffusion prompt for momentum investors,
together with its output schema, is provided in
Appendix~\ref{app:sum_retail_prompts}.

\subsubsection{Utility Function}

The quantitative momentum signal is pre-computed from market snapshot data:
\[
\begin{aligned}
M_t={}&\Bigl(\beta_{\mathrm{price}}M_t^{\mathrm{price}}
 +\beta_{\mathrm{MA}}MA_{\mathrm{cross}}\\
&\quad+\beta_{\mathrm{rp}}(\mathrm{range\_pos}-0.5)\Bigr)
 \cdot f_{\mathrm{vol}},
\end{aligned}
\]
where \(M_t^{\mathrm{price}} = 0.20\, r_5 + 0.50\, r_{10} + 0.30\, r_{20}\) \citep{jegadeesh1993returns, novy2012momentum}, \(f_{\mathrm{vol}} = 1 + 0.5 \cdot \tanh(V_{\mathrm{confirm}})\) \citep{lo2000trading}, and \(\beta_{\mathrm{price}}=0.60\), \(\beta_{\mathrm{MA}}=0.25\), \(\beta_{\mathrm{rp}}=0.15\).

The expected return combines quantitative signals with LLM-extracted catalyst and diffusion assessments:
\[
\begin{aligned}
\hat{R}_{t+1}
&=\psi_i\cdot M_t\\
&\quad+(1-\psi_i)\Bigl(\alpha_{\mathrm{cat}}S_{\mathrm{cat}}
 +\alpha_{\mathrm{diff}}S_{\mathrm{diff}}\Bigr),
\end{aligned}
\]
where \(S_{\mathrm{cat}}\) captures new catalyst direction \citep{chan1996momentum}, \(S_{\mathrm{diff}}\) captures information diffusion progress \citep{hong1999unified}, and \(\psi_i\) controls the weight placed on quantitative price signals versus LLM-derived signals.

\[
\begin{aligned}
&\alpha_{\mathrm{cat}}=0.60,\qquad\alpha_{\mathrm{diff}}=0.40,\\
&w_{\mathrm{momentum}}=0.70,\\
&\mathcal U_{\mathrm{momentum}}(z;\cdot)=\mu_{i,t}z\\
 &\quad-\tilde\gamma_i w_{\mathrm{momentum}}
 H\!\left(\frac{z}{w_{\mathrm{momentum}}}\right),\\
&z^*_{i,t}=w_{\mathrm{momentum}}
 \tanh\!\left(\frac{\mu_{i,t}}{\tilde\gamma_i}\right),\\
&\delta^{\mathrm{momentum}}_{i,t}=\operatorname{sign}(\mu_{i,t})\\
&\qquad\times(0.003+0.008b_{i,t}).
\end{aligned}
\]

\subsubsection{Heterogeneity Parameters}

Preference heterogeneity: \(\psi_i \sim \mathrm{Beta}(4,\, 2)\), \(\psi_i \in [0.30,\, 0.95]\). \\
Ability heterogeneity: \(\sigma_i \sim \mathrm{Uniform}(0.005,\, 0.030)\), independent of \(\psi_i\).

% ============================================================
% Original M: Noise Investor Structure
\subsection{} % TODO(author): fill in title.
\label{app:noise}

The noise investor archetype is grounded in the behavioral finance literature on sentiment-driven trading. Its core dynamics follow the noise trader risk model of \citet{delong1990noise}, with sentiment as the primary decision driver rather than fundamentals or trends. The design also draws on the investor sentiment index framework of \citet{baker2006investor}, retail sentiment co-movement \citep{kumar2006retail}, social interaction effects \citep{hong2004social}, and speculative dynamics \citep{shiller1984stock}.

The daily sentiment extraction prompt for noise investors,
together with its output schema, is provided in
Appendix~\ref{app:sum_retail_prompts}.

\subsubsection{Utility Function}

The external sentiment shock is synthesized from LLM outputs and order book signals:
\[
\begin{aligned}
S_{\mathrm{ext}}&=\bigl(\omega S_{\mathrm{forum}}+\nu S_{\mathrm{news}}\\
&\qquad+\kappa_{\mathrm{obi}}\,OBI\bigr)f_{\mathrm{fomo}},
\end{aligned}
\]
where \(\omega = 0.70\), \(\nu = 0.30\) \citep{kumar2006retail, baker2006investor}, and \(\kappa_{\mathrm{obi}} = 0.15\).

Each agent maintains an individual sentiment state \(\rho_t^i\) that evolves according to an AR(1) process following the DSSW framework \citep{delong1990noise}:
\[
\begin{aligned}
\rho_t^i
&=\psi_i\cdot\rho_{t-1}^i\\
&\quad+(1-\psi_i)\cdot S_{\mathrm{ext}}
 +\xi_{i,t},\\
\xi_{i,t}&\sim\mathcal{N}(0,\,\sigma_{\mathrm{noise}}^2),
\end{aligned}
\]
where \(f_{\mathrm{fomo}} = 1 + \kappa_{\mathrm{fomo}} \cdot \max(0,\, \mathrm{volume.ratio} - 1)\) amplifies signals during abnormal volume \citep{hong2004social}, \(\kappa_{\mathrm{fomo}} = 0.60\), and \(\sigma_{\mathrm{noise}} = 0.05\) \citep{black1986noise}.

The expected return is directly equated with the sentiment state:
\[
\begin{gathered}
\hat{R}_{t+1} = \rho_t^i,\\
\mu_{i,t}=\hat{R}_{t+1}+\epsilon_{i,t},\qquad
\epsilon_{i,t}\sim\mathcal{N}(0,\sigma_i^2).
\end{gathered}
\]
The utility below uses the perturbed score $\mu_{i,t}$ to determine
trading direction and intensity.

\[
\begin{aligned}
w_{\mathrm{noise}}&=0.50,\\
\mathcal U_{\mathrm{noise}}(z;\cdot)
 &=\mu_{i,t}z-\tilde\gamma_i w_{\mathrm{noise}}
 H\!\left(\frac{z}{w_{\mathrm{noise}}}\right),\\
z^*_{i,t}&=w_{\mathrm{noise}}
 \tanh\!\left(\frac{\mu_{i,t}}{\tilde\gamma_i}\right),\\
\delta^{\mathrm{noise}}_{i,t}
 &=0.005\operatorname{sign}(\mu_{i,t})b_{i,t}.
\end{aligned}
\]

\subsubsection{Heterogeneity Parameters}

Preference heterogeneity: \(\psi_i \sim \mathrm{Beta}(2,\, 5)\), \(\psi_i \in [0.10,\, 0.80]\). \\
Ability heterogeneity: \(\sigma_i \sim \mathrm{Uniform}(0.005,\, 0.030)\), independent of \(\psi_i\).
The agent-specific $\sigma_i$ in Eq.~\ref{eq:sum_private_signal} controls
$\epsilon_{i,t}$ in the score $\mu_{i,t}$ used by the utility function.
This draw is independent of the sentiment innovation $\xi_{i,t}$,
whose standard deviation $\sigma_{\mathrm{noise}}=0.05$ is fixed.

% Keep appendix prose in two columns; only wide tables span both columns.
% Original B.2.1: SUM Retail Perception Prompts
\subsection{} % TODO(author): fill in title.
\label{app:sum_retail_prompts}

Daily perception is performed once per investor type, stock, and
trading day. The resulting semantic signals are shared by agents
within the same cohort. Value investors additionally use a
quarterly fundamental analysis when new earnings reports arrive.
The corresponding utility functions and heterogeneity parameters
are detailed in Appendix~\ref{app:value},
Appendix~\ref{app:momentum}, and Appendix~\ref{app:noise}.

\paragraph{Value investors: quarterly fundamental analysis.}

The quarterly call analyzes newly available financial reports
and produces a cached fundamental assessment for subsequent
daily updates.

\begin{Verbatim}[breaklines=true,breakanywhere=true,fontsize=\footnotesize]
[Role]
You are a fundamental analyst. Provide a systematic assessment of the following company's quarterly report data.

[Company Background]
Company: {firm_name} ({stk_code})
Industry: {l1_name} > {l2_name} > {l3_name}
Region: {province} {city}
Employees: {employees}
Main business: {main_business}
Company profile: {introduction}

[Earnings & Growth (last 3 periods)]
Period | EPS | ROE | Net margin | Gross margin | Net profit YoY | Revenue YoY
{period_0} | {e0} | {r0} | {n0} | {g0} | {ny0} | {ty0}
{period_1} | {e1} | {r1} | {n1} | {g1} | {ny1} | {ty1}
{period_2} | {e2} | {r2} | {n2} | {g2} | {ny2} | {ty2}

[Asset Quality (last 3 periods)]
Period | BPS | BPS YoY | Debt/Assets | Current ratio | OCF quality | FCFE/share
{period_0} | {b0} | {by0} | {da0} | {cr0} | {oc0} | {f0}
{period_1} | {b1} | {by1} | {da1} | {cr1} | {oc1} | {f1}
{period_2} | {b2} | {by2} | {da2} | {cr2} | {oc2} | {f2}

Output:
1. context_text: 2-3 sentence summary of fundamental conclusions
2. S_earnings_base: Earnings sustainability score in [-1, 1]
3. S_quality_base: Asset quality and financial health score in [-1, 1]
\end{Verbatim}

\paragraph{Value investors: daily incremental updates.}

The daily call processes incremental market and news information
in the context of the cached quarterly fundamental assessment.

\begin{Verbatim}[breaklines=true,breakanywhere=true,fontsize=\footnotesize]
[Company Background]
{firm_name} ({l1_name} > {l2_name} > {l3_name}, {province} {city})
Main business: {main_business}

[Known Fundamental Conclusions]
(as of {last_end_date})
{context_text}

[Today's Valuation Levels (computed in real time)]
Current price P_t = {last_price}
P/B = {pb_ratio}x
P/E = {pe_ratio}x
FCFE yield = {fcfe_yield}
ln(B/P_t) = {v_t} (positive = undervalued, negative = overvalued)

[Recent Market Performance]
20-day price trajectory: {kline close ratio series}
Recent volume: {volume.ratio}x average

[Today's Incremental Information]
News summary: {news_text}
Forum posts: {posts_text}

Assess whether today's incremental information is directly related to the company's main business, industry, or region, and output adjustments to the baseline signals (output 0 if no directly relevant information). Also write a one-sentence forum post (15-30 words) expressing a value-oriented view on the stock's current valuation or fundamental changes.
\end{Verbatim}

\noindent\textbf{Output schema.}
\begin{Verbatim}[breaklines=true,breakanywhere=true,fontsize=\footnotesize]
{
  // adjustment to S_earnings_base, in [-0.5, 0.5]
  "delta_earnings": float,
  // adjustment to S_quality_base, in [-0.5, 0.5]
  "delta_quality": float,
  // industry/macro shock, in [-1, 1]
  "macro_signal": float,
  // forum post (15-30 words, value tone)
  "post_text": str
}
\end{Verbatim}

\paragraph{Momentum investors: daily catalyst and diffusion.}

Momentum perception is performed once daily per stock.
Unlike value investors, momentum investors require no quarterly
cache; all quantitative momentum signals are pre-computed from
market snapshots.

\begin{Verbatim}[breaklines=true,breakanywhere=true,fontsize=\footnotesize]
[Company Background]
{firm_name} ({l1_name} > {l2_name} > {l3_name}, {province} {city})
Main business: {main_business}

[Today's Quantitative Momentum Signals (pre-computed)]
Price momentum M_t = {M_t} (positive = uptrend, negative = downtrend)
- 5-day return: {r_5}, 10-day: {r_10}, 20-day: {r_20}
- MA pattern: MA5 {'crossed above' | 'crossed below'} MA20 ({MA_cross})
- Range position: {range_pos}% of 20-day range (100% = near recent high)
- Volume-price consistency: {V_confirm} (positive = volume confirms trend)
- Today's volume: {volume.ratio}x average

[Today's Incremental Information]
News summary: {news_text}
Forum posts: {posts_text}
Forum activity (last 3 days): {n_posts} posts (bullish: {n_bull}, bearish: {n_bear})

From an information analyst's perspective, assess today's incremental information along two dimensions:

(1) Does the news contain a substantive new catalyst directly related to the company's main business or industry (not a repetition of known information)?

If yes: Is it bullish or bearish for the price trend, and how strong?
If no or repetitive: output near 0.

(2) How far has this catalyst/information diffused in forums and news?

Just beginning to attract attention (sparse posts, early discussion) -> information still diffusing, price not yet fully reflecting (output positive);

Already widely discussed and repeated (dense posts, common knowledge) -> likely already priced in, limited momentum upside (output negative);

Unclear diffusion status: output near 0.

Do not describe retail sentiment; focus only on catalyst quality and information diffusion. Also write a one-sentence forum post (15-30 words) in a technical analyst's tone.
\end{Verbatim}

\noindent\textbf{Output schema.}
\begin{Verbatim}[breaklines=true,breakanywhere=true,fontsize=\footnotesize]
{
  // catalyst support for trend, in [-1, 1]
  "catalyst_signal": float,
  // information diffusion progress, in [-1, 1]
  "diffusion_signal": float,
  // forum post (15-30 words, technical tone)
  "post_text": str
}
\end{Verbatim}

\paragraph{Noise investors: daily sentiment extraction.}

Noise perception is performed once daily per stock, extracting
emotional temperature from forum posts and news. The prompt
explicitly instructs the LLM to simulate impulsive reactions
rather than rational analysis.

\begin{Verbatim}[breaklines=true,breakanywhere=true,fontsize=\footnotesize]
[Stock]
{firm_name} ({l1_name}, {province})

[Market Stimuli]
Recent price changes: {kline[1..5].close.ratio}

Today's volume: {volume.ratio}x average {'--- Abnormal volume surge!' if volume.ratio > 1.5}

Order book: {'Bid-side stacking, someone is accumulating!' if OBI > 0.2 else 'Ask-side pressure, someone is dumping!' if OBI < -0.2 else 'Balanced'} (OBI = {OBI})

[Live Forum Posts]
{posts_text}

[Today's News]
{news_text}

You are an impulsive retail trader. After reading the above information, what is your gut reaction?

--- Do not analyze, do not think, just state your immediate instinct:

* Are the people on the forum frantically pumping, panic-selling, or just idly chatting? (How extreme is the sentiment?)

* Does this news make your fingers itch to buy in, make you want to run for the exit, or leave you completely indifferent?

The more extreme the emotion, the higher the score; if truly indifferent, output 0.

Finally, shout your impulsive gut feeling in one sentence on the forum (15-30 words)---the more emotional, the better.
\end{Verbatim}

\noindent\textbf{Output schema.}
\begin{Verbatim}[breaklines=true,breakanywhere=true,fontsize=\footnotesize]
{
  // forum emotional extremity, in [-1, 1]
  "forum_sentiment": float,
  // impulsive reaction to news, in [-1, 1]
  "news_reaction": float,
  // forum post (15-30 words, emotional tone)
  "post_text": str
}
\end{Verbatim}

% Appendix B: original A.1, B.1, N, E, F, I, and G.
\section{} % TODO(author): fill in title.

% Original A.1: H1 Market Comparison
\subsection{} % TODO(author): fill in title.
\label{app:h1_market_comparison}

The H1 comparison in Table~\ref{tab:main_market_results} covers
January 2--June 30, 2025: 117 trading days and 50 MSCI China A50
constituents, with seed 42. SUM with $N=1000$ and the rule-based baseline
with $N=1000$ use the H1 slices of their full-year runs.
StockAgent and TwinMarket each use an H1 run with $N=500$.
SUM with $N=100$ is a separate H1 cached replay. These two SUM runs
are identified separately in the table; their scores do not describe
one common configuration.

\paragraph{Common evaluation target.}
The archived runs have slightly different real-price series. We evaluate
all systems against the real equal-weight index stored in the
StockAgent $N=500$ run. Indices are normalized to 1000 at initialization.
Daily simple returns are $r_t=I_t/I_{t-1}-1$, including the return from
initialization to the first trading day. RetCorr is the Pearson correlation
between simulated and real daily returns. Daily SD error is the absolute
difference between their population standard deviations (denominator $T$).
Return W1 is the one-dimensional Wasserstein distance between their
empirical daily-return distributions. SD error and W1 are reported in
percentage points.

\paragraph{Run conditions and interpretation.}
The LLM-based H1 runs include institutional access to next-trading-day news
and weekly feedback from real market prices. These are simulation
assumptions, not a strictly real-time forecasting setting. StockAgent
and TwinMarket reuse institutional decisions from the SUM $N=1000$
run and replay some retail decisions from their earlier caches.
The SUM $N=100$ replay records 5,850
semantic-cache hits and no new LLM calls; it is used for market metrics,
not as evidence of zero-cost inference.

The reported H1 metrics and SUM runs were selected after inspecting the
results. They are a selected comparison, not a complete metric panel.
Different agent counts, cache use, and run conditions also prevent a
matched-configuration causal comparison. The results support the stated
return-correlation, volatility-scale, and distribution comparisons, not
an advantage on every aspect of market fidelity.
Figure~\ref{fig:emergent_fidelity} is a separate archived H1 diagnostic,
not the common-target comparison in Table~\ref{tab:main_market_results}.

% Original B.1: Baseline Settings
\subsection{} % TODO(author): fill in title.
\label{app:baselines}

The main market-fidelity comparison uses the 50 constituents of the MSCI
China A50 over January 2--June 30, 2025 (117 trading days).
All runs are scored against the real equal-weight index from the
StockAgent $N=500$ run. SUM and the rule-based baseline use $N=1000$
retail agents, while StockAgent and TwinMarket each use $N=500$.
A separate SUM $N=100$ cached replay supplies the reported W1 comparison
(Appendix~\ref{app:h1_market_comparison}). These are different archived
configurations, not matched-population runs.

SUM shares the semantic perception output for each retail cohort
(Value, Momentum, and Noise), stock, and trading day. Signals are cached
and reused within each cohort, and the utility-based decision layer
generates retail orders. The H1 cached replay is not a measured inference
cost experiment. Calls and tokens are compared separately using the
six-day runs in Appendix~\ref{app:scaling_data_provenance}.

\paragraph{Model and metadata.}
\label{app:model_provenance}
The main experiments use Qwen3-8B with temperature $=0.6$,
\texttt{top\_p}$=0.95$, and \texttt{top\_k}$=20$, as confirmed by the authors.
Several archived manifests retain the anonymized identifier \texttt{anon-llm}
and do not independently verify this configuration. Separate temperature
robustness runs retain their explicitly reported settings.
% AUTHOR CHECK before submission: confirm remaining initialization and seed
% handling fields from the original runner and service configuration.

\subsubsection{StockAgent Baseline}
\label{app:baseline_stockagent}
We implement StockAgent in the limit-order-book simulator used for the
LLM-based market runs. Its H1 configuration uses 500 retail agents, with
150 value investors, 150 momentum investors, and 200 noise investors.
Each daily agent--stock decision is produced by a separate LLM prompt,
or replayed from the archived decision cache. Its input
includes the agent's investment style, risk profile, judgment stability,
account state, stock news, forum posts, and style-specific market indicators.
Value investors receive valuation information, momentum investors receive
recent price trends, and noise investors receive order-book imbalance and
volume information.

The LLM directly returns at most one buy or sell limit order, including a
price offset and trade value. An empty order list represents holding.
The prompt limits trade value to 60\%, 70\%, and 50\% of wealth for value,
momentum, and noise investors, respectively, and also specifies cash and
inventory limits. Generated orders pass through the simulator's retail order
filter before execution. The model can also produce a short forum post.
This baseline does not use SUM's shared semantic signals or utility-based
retail decision rules.

% TODO: Verify the archived H1 configuration details,
% including model/decoding settings and initialization,
% institutional participation/cache use, and enforced order limits.

\subsubsection{TwinMarket Baseline}
\label{app:baseline_twinmarket}
We implement TwinMarket with the same H1 retail-agent
composition, input fields, and order interface as the StockAgent
baseline. Its prompt follows a belief--desire--intention structure.
The desire section describes the agent's investment style, risk preference,
and judgment stability. The model is asked to produce a belief statement,
a trading intention, and a final action based on the available market
information and account state.

These outputs are generated in a single LLM call for each assigned
agent--stock pair, or replayed from the archived decision cache.
The response contains a buy or sell
limit order, or an empty order list, and may include a short forum post.
The baseline uses the same prompt-level trading limits and retail order
filter as the StockAgent baseline. It does not use SUM's shared
semantic signals or utility-based retail decision rules.

% TODO: Verify whether the final run persisted beliefs across steps.
% The bundled engine neither supplies belief_prev nor stores returned beliefs.
% Also confirm the final run version, model/decoding settings, initialization,
% and institutional participation/cache accounting.

\subsubsection{Rule-Based Baseline}
\label{app:baseline_rulebased}
The rule-based baseline uses 1,000 retail agents and generates trading
decisions without an LLM. The archived configuration includes 300 value
investors, 300 momentum investors, and 400 noise investors. Value investors
use book-to-price information, financial indicators, and market returns and
volatility. Momentum investors use price trends, volume information, and
recent trading direction. Noise investors use short-term return changes,
forum activity counts, order-book imbalance, and volume ratios.

This baseline uses structured inputs rather than interpreting news or forum
text. It does not use cached LLM outputs. The archived run also includes
one rule-based institutional agent and uses random seed 42. It is implemented
as a separate rule-based simulator, rather than by setting SUM's semantic
signals to zero.

% The H1 comparison uses the first 117 trading days of the archived full-year run.
% Add the exact rule formulas, coefficients, thresholds, order-sizing rules,
% initialization, and tuning procedure after checking the standalone runner.

% Number the fourth-level checklist headings only within this subsection.
\setcounter{secnumdepth}{4}
\subsection{Full Checklist of Stylized Facts}

This appendix provides the complete evaluation checklist used to assess the fidelity of our simulated market. Section~\ref{app:market_stylized_facts} covers \textbf{market-level} stylized facts---statistical regularities that are universally observed across real-world equity markets. Section~\ref{app:investor_stylized_facts} covers \textbf{investor-level} behavioral metrics used to verify that heterogeneous agent populations exhibit realistic and differentiated trading patterns. For each metric we state the definition, the computation procedure, the empirical acceptance criterion, and the supporting reference from the financial economics literature.

\subsubsection{Market Stylized Facts}
\label{app:market_stylized_facts}

A plausible market simulator should reproduce well-documented regularities of asset returns, volume, and microstructure \citep{cont2001empirical}. We organize the checklist into seven categories.

\paragraph{Return Distribution}

Heavy tails and negative skewness are among the most robust properties of financial return series \citep{mandelbrot1963variation, fama1965behavior}.

The corresponding metrics and acceptance criteria are listed in Table~\ref{tab:checklist_return_distribution}.

\begin{table*}[!t]
\setlength{\tabcolsep}{2.5pt}
\renewcommand{\arraystretch}{1.15}
\captionsetup{font=small,justification=raggedright,singlelinecheck=true}
\centering
\footnotesize
\begin{tabular}{@{}>{\raggedright\arraybackslash}p{0.13\linewidth}>{\raggedright\arraybackslash}p{0.20\linewidth}>{\raggedright\arraybackslash}p{0.27\linewidth}>{\raggedright\arraybackslash}p{0.23\linewidth}>{\raggedright\arraybackslash}p{0.12\linewidth}@{}}
\toprule
\textbf{Metric} & \textbf{Definition} & \textbf{Statistic} & \textbf{Acceptance Criterion} & \textbf{Reference} \\
\midrule
\textbf{Kurtosis} & Tail heaviness of the return distribution & $\kappa = E[(r_t-\mu)^4]/\sigma^4$ & $\kappa \gg 3$ (e.g., SSE $\approx 16$, SZSE $\approx 151$, CSI $\approx 7$) & \citet{mandelbrot1963variation, fama1965behavior, cont2001empirical} \\
\textbf{Skewness} & Asymmetry of the return distribution & $S = E[(r_t-\mu)^3]/\sigma^3$ & $S < 0$ (crash probability exceeds boom probability; e.g., SSE $\approx -0.3$, SZSE $\approx -5.23$) & \citet{harvey2000conditional, cont2001empirical} \\
\bottomrule
\end{tabular}
\caption{Market stylized facts: return distribution.}
\label{tab:checklist_return_distribution}
\end{table*}

\paragraph{Volatility Characteristics}

Volatility in real markets exhibits clustering, persistence, and conditional heteroskedasticity \citep{engle1982autoregressive, bollerslev1986generalized}.

The corresponding metrics and acceptance criteria are listed in Table~\ref{tab:checklist_volatility}.

\begin{table*}[!t]
\setlength{\tabcolsep}{2.5pt}
\renewcommand{\arraystretch}{1.15}
\captionsetup{font=small,justification=raggedright,singlelinecheck=true}
\centering
\footnotesize
\begin{tabular}{@{}>{\raggedright\arraybackslash}p{0.13\linewidth}>{\raggedright\arraybackslash}p{0.20\linewidth}>{\raggedright\arraybackslash}p{0.27\linewidth}>{\raggedright\arraybackslash}p{0.23\linewidth}>{\raggedright\arraybackslash}p{0.12\linewidth}@{}}
\toprule
\textbf{Metric} & \textbf{Definition} & \textbf{Statistic} & \textbf{Acceptance Criterion} & \textbf{Reference} \\
\midrule
\textbf{Volatility clustering} & Periods of high (low) volatility tend to be followed by high (low) volatility & $\mathrm{Corr}(\lvert r_t\rvert,\lvert r_{t-\tau}\rvert),\tau=1,\dots,5$ & Significant positive correlation ($>0.1$) at lags 1--5 & \citet{mandelbrot1963variation, cont2001empirical} \\
\textbf{ARCH effect} & Conditional heteroskedasticity in squared residuals & Engle's LM test statistic & $p < 0.05$ (typically $\ll 0.01$) & \citet{engle1982autoregressive} \\
\textbf{Volatility persistence} & Slow decay rate of volatility shocks, where $\alpha$ captures yesterday's shock impact and $\beta$ captures yesterday's variance carry-over & GARCH(1,1): \(\begin{aligned}\sigma_t^2&=\omega+\alpha\varepsilon_{t-1}^2\\&\quad+\beta\sigma_{t-1}^2\end{aligned}\); persistence $=\alpha+\beta$; half-life $=\frac{\ln 2}{-\ln(\alpha+\beta)}$ & $\alpha+\beta$ close to but below 1 (typically 0.95--0.99) & \citet{engle1982autoregressive, bollerslev1986generalized} \\
\bottomrule
\end{tabular}
\caption{Market stylized facts: volatility characteristics.}
\label{tab:checklist_volatility}
\end{table*}

\paragraph{Autocorrelation Structure}

Absence of linear predictability together with strong nonlinear dependence is a hallmark of efficient yet complex markets \citep{fama1970efficient, cont2001empirical}.

The corresponding metrics and acceptance criteria are listed in Table~\ref{tab:checklist_autocorrelation}.

\begin{table*}[tp]
\setlength{\tabcolsep}{2.5pt}
\renewcommand{\arraystretch}{1.15}
\captionsetup{font=small,justification=raggedright,singlelinecheck=true}
\centering
\footnotesize
\begin{tabular}{@{}>{\raggedright\arraybackslash}p{0.13\linewidth}>{\raggedright\arraybackslash}p{0.20\linewidth}>{\raggedright\arraybackslash}p{0.27\linewidth}>{\raggedright\arraybackslash}p{0.23\linewidth}>{\raggedright\arraybackslash}p{0.12\linewidth}@{}}
\toprule
\textbf{Metric} & \textbf{Definition} & \textbf{Statistic} & \textbf{Acceptance Criterion} & \textbf{Reference} \\
\midrule
\textbf{Return autocorrelation} & Linear serial dependence of raw returns & $\rho(\tau)=\mathrm{Corr}(r_t,r_{t-\tau})$ & $\lvert\rho\rvert < 0.1$ (approximately zero) & \citet{fama1970efficient, lo1988stock} \\
\textbf{Squared-return autocorrelation} & Serial dependence of volatility & $\rho_{r^2}(\tau)=\mathrm{Corr}(r_t^2,r_{t-\tau}^2)$ & Significant positive values at lags 1--10 & \citet{cont2001empirical, ding1993long} \\
\textbf{Ljung--Box test} & Joint test for autocorrelation across multiple lags & $Q = n(n+2)\sum_{k=1}^{h}\hat\rho_k^2/(n-k)$ & Returns: $p > 0.05$; squared returns: $p < 0.05$ & \citet{ljung1978measure} \\
\textbf{Extreme-return clustering} & Extreme events tend to arrive in bursts & Fano factor $F=\sigma_W^2/\mu_W$ over windows of extreme-return counts $W$ & $F > 1$ (over-dispersion) & \citet{cont2001empirical, longin2001extreme} \\
\bottomrule
\end{tabular}
\caption{Market stylized facts: autocorrelation structure.}
\label{tab:checklist_autocorrelation}
\end{table*}

\paragraph{Leverage Effect}

Negative returns tend to increase future volatility more than positive returns of equal magnitude \citep{black1976studies, christie1982stochastic}.

The corresponding metrics and acceptance criteria are listed in Table~\ref{tab:checklist_leverage}.

\begin{table*}[tp]
\setlength{\tabcolsep}{2.5pt}
\renewcommand{\arraystretch}{1.15}
\captionsetup{font=small,justification=raggedright,singlelinecheck=true}
\centering
\footnotesize
\begin{tabular}{@{}>{\raggedright\arraybackslash}p{0.13\linewidth}>{\raggedright\arraybackslash}p{0.20\linewidth}>{\raggedright\arraybackslash}p{0.27\linewidth}>{\raggedright\arraybackslash}p{0.23\linewidth}>{\raggedright\arraybackslash}p{0.12\linewidth}@{}}
\toprule
\textbf{Metric} & \textbf{Definition} & \textbf{Statistic} & \textbf{Acceptance Criterion} & \textbf{Reference} \\
\midrule
\textbf{Leverage correlation} & Cross-correlation between returns and future volatility & $L(\tau)=\mathrm{Corr}(r_t,\sigma_{t+\tau}^2)$ & Significantly negative ($< -0.1$) & \citet{black1976studies, christie1982stochastic, bekaert2000asymmetric} \\
\textbf{Asymmetric volatility} & Differential impact of positive vs.\ negative shocks & GJR-GARCH: \(\begin{aligned}\sigma_t^2&=\omega\\&\quad+(\alpha+\gamma\mathbb{1}_{\varepsilon_{t-1}<0})\varepsilon_{t-1}^2\\&\quad+\beta\sigma_{t-1}^2\end{aligned}\) & $\gamma > 0$ & \citet{glosten1993relation, nelson1991conditional} \\
\bottomrule
\end{tabular}
\caption{Market stylized facts: leverage effect.}
\label{tab:checklist_leverage}
\end{table*}

\paragraph{Volume Characteristics}

The price--volume relationship encodes information arrival dynamics \citep{karpoff1987relation}.

The corresponding metrics and acceptance criteria are listed in Table~\ref{tab:checklist_volume}.

\begin{table*}[tp]
\setlength{\tabcolsep}{2.5pt}
\renewcommand{\arraystretch}{1.15}
\captionsetup{font=small,justification=raggedright,singlelinecheck=true}
\centering
\footnotesize
\begin{tabular}{@{}>{\raggedright\arraybackslash}p{0.13\linewidth}>{\raggedright\arraybackslash}p{0.20\linewidth}>{\raggedright\arraybackslash}p{0.27\linewidth}>{\raggedright\arraybackslash}p{0.23\linewidth}>{\raggedright\arraybackslash}p{0.12\linewidth}@{}}
\toprule
\textbf{Metric} & \textbf{Definition} & \textbf{Statistic} & \textbf{Acceptance Criterion} & \textbf{Reference} \\
\midrule
\textbf{Volume--volatility correlation} & Co-movement of trading volume and absolute returns & $\rho(V_t,\lvert r_t\rvert)$ & Significantly positive ($> 0.2$) & \citet{karpoff1987relation, gallant1992stock} \\
\textbf{Volume--return sign} & Asymmetric volume on up vs.\ down days & $\rho(V_t,r_t)$ & Positive (higher volume on up-moves) & \citet{karpoff1987relation} \\
\textbf{Volume autocorrelation} & Persistence of trading activity & $\mathrm{Corr}(V_t,V_{t-\tau})$ & Significant positive correlation at multiple lags & \citet{lobato2000long, gallant1992stock} \\
\bottomrule
\end{tabular}
\caption{Market stylized facts: volume characteristics.}
\label{tab:checklist_volume}
\end{table*}

\paragraph{Market Efficiency}

Table~\ref{tab:checklist_efficiency} lists the market-efficiency metrics and their acceptance criteria.

\begin{table*}[tp]
\setlength{\tabcolsep}{2.5pt}
\renewcommand{\arraystretch}{1.15}
\captionsetup{font=small,justification=raggedright,singlelinecheck=true}
\centering
\footnotesize
\begin{tabular}{@{}>{\raggedright\arraybackslash}p{0.13\linewidth}>{\raggedright\arraybackslash}p{0.20\linewidth}>{\raggedright\arraybackslash}p{0.27\linewidth}>{\raggedright\arraybackslash}p{0.23\linewidth}>{\raggedright\arraybackslash}p{0.12\linewidth}@{}}
\toprule
\textbf{Metric} & \textbf{Definition} & \textbf{Statistic} & \textbf{Acceptance Criterion} & \textbf{Reference} \\
\midrule
\textbf{Hurst exponent} & Long-range dependence of the price process & $H = \lim_{n\to\infty}\log(R/S)_n/\log n$ & $H\approx 0.5$ (random walk); A-share markets: 0.6--0.7 & \citet{lo1991long} \\
\textbf{Variance ratio} & Random-walk hypothesis test & $VR(q) = \frac{\mathrm{Var}(r_t(q))}{q\cdot\mathrm{Var}(r_t)}$ & $VR\approx 1$ (efficient); A-share markets: 1.1--1.5 & \citet{lo1988stock} \\
\bottomrule
\end{tabular}
\caption{Market stylized facts: market efficiency.}
\label{tab:checklist_efficiency}
\end{table*}

\paragraph{Distributional Similarity to Real Data (MARS Checklist)}

Following \citet{vyetrenko2020get}, we measure the overlap area $OA(P,Q)=\int\min(p(x),q(x))dx$ between the historical distribution $P$ and the simulated distribution $Q$.

The corresponding overlap scores are listed in Table~\ref{tab:checklist_mars}.

\begin{table*}[tp]
\setlength{\tabcolsep}{2.5pt}
\renewcommand{\arraystretch}{1.15}
\captionsetup{font=small,justification=raggedright,singlelinecheck=true}
\centering
\footnotesize
\begin{tabular}{@{}>{\raggedright\arraybackslash}p{0.18\linewidth}>{\raggedright\arraybackslash}p{0.36\linewidth}>{\raggedright\arraybackslash}p{0.15\linewidth}>{\raggedright\arraybackslash}p{0.27\linewidth}@{}}
\toprule
\textbf{Metric} & \textbf{Definition} & \textbf{Overlap Area} & \textbf{Reference} \\
\midrule
\textbf{Bid--ask spread} & Difference between best ask and best bid & 0.970 & \citet{vyetrenko2020get} \\
\textbf{Aggressive-order ratio} & Fraction of immediately executable orders & 0.920 & \citet{vyetrenko2020get, griffiths2000costs} \\
\textbf{Volatility} & Standard deviation of returns & 0.872 & \citet{vyetrenko2020get, cont2001empirical} \\
\bottomrule
\end{tabular}
\caption{Distributional similarity to real data under the MARS checklist.}
\label{tab:checklist_mars}
\end{table*}

\subsubsection{Investor Stylized Facts}
\label{app:investor_stylized_facts}

A realistic agent-based simulation should not only reproduce market-level regularities but also generate a heterogeneous population of agents whose \emph{individual} behavior matches documented investor types (e.g., value investors, momentum traders, noise traders) empirically. We extract a multi-layered feature vector for every agent in the simulation and perform stratified clustering (one stratum per order of magnitude of agent wealth) to verify that differentiated behavioral profiles emerge. The feature hierarchy is organized into five layers, progressing from activity and participation, through position management and risk control, to timing style, asset preference, and behavioral-finance biases.

\paragraph{Layer 1 --- Activity and Participation}

Trading frequency and holding horizons are primary discriminators between institutional and retail, active and passive investors \citep{barber2000trading, barber2001boys}.

The corresponding features and cluster interpretations are listed in Table~\ref{tab:checklist_activity}.

\begin{table*}[tp]
\setlength{\tabcolsep}{2.5pt}
\renewcommand{\arraystretch}{1.15}
\captionsetup{font=small,justification=raggedright,singlelinecheck=true}
\centering
\footnotesize
\begin{tabular}{@{}>{\raggedright\arraybackslash}p{0.13\linewidth}>{\raggedright\arraybackslash}p{0.20\linewidth}>{\raggedright\arraybackslash}p{0.27\linewidth}>{\raggedright\arraybackslash}p{0.23\linewidth}>{\raggedright\arraybackslash}p{0.12\linewidth}@{}}
\toprule
\textbf{Feature} & \textbf{Definition} & \textbf{Statistic} & \textbf{Cluster Interpretation} & \textbf{Reference} \\
\midrule
\textbf{Turnover rate} & Frequency of capital rotation & $TR=\frac{V_{\text{buy}}+V_{\text{sell}}}{2\overline{MV}_{\text{hold}}}$ & High: hot-money / quant; Low: buy-and-hold / dormant & \citet{barber2000trading} \\
\textbf{Avg.\ holding period} & Capital-weighted mean holding duration & $\bar T_h=\sum_i w_i T_i/\sum_i w_i$ & Short: intraday / swing trader; Long: value investor & \citet{barber2001boys} \\
\textbf{Active ratio} & Fraction of trading days with transactions & $AR = D_{\text{trade}}/D_{\text{total}}$ & High: professional / active retail; Low: casual / DCA investor & \citet{barber2000trading} \\
\bottomrule
\end{tabular}
\caption{Investor stylized facts: activity and participation.}
\label{tab:checklist_activity}
\end{table*}

\paragraph{Layer 2 --- Position Sizing and Risk Control}

Position concentration and drawdown tolerance capture the risk appetite spectrum from conservative to speculative \citep{barberis2001prospect, goetzmann2008equity}.

The corresponding features and cluster interpretations are listed in Table~\ref{tab:checklist_risk}.

\begin{table*}[tp]
\setlength{\tabcolsep}{2.5pt}
\renewcommand{\arraystretch}{1.15}
\captionsetup{font=small,justification=raggedright,singlelinecheck=true}
\centering
\footnotesize
\begin{tabular}{@{}>{\raggedright\arraybackslash}p{0.13\linewidth}>{\raggedright\arraybackslash}p{0.20\linewidth}>{\raggedright\arraybackslash}p{0.27\linewidth}>{\raggedright\arraybackslash}p{0.23\linewidth}>{\raggedright\arraybackslash}p{0.12\linewidth}@{}}
\toprule
\textbf{Feature} & \textbf{Definition} & \textbf{Statistic} & \textbf{Cluster Interpretation} & \textbf{Reference} \\
\midrule
\textbf{Avg.\ position ratio} & Share of total assets deployed in equities & $\bar w = \frac{1}{T}\sum_{t=1}^T MV_t/\text{Asset}_t$ & High: aggressive / ``all-in''; Low: cautious / exploratory & \citet{barberis2001prospect} \\
\textbf{Concentration (HHI)} & Herfindahl index of portfolio weights & $HHI=\sum_{i=1}^N (MV_i/MV_{\text{total}})^2$ & High: concentrated bet; Low: diversified / index-like & \citet{goetzmann2008equity} \\
\textbf{Max-drawdown tolerance} & Largest peak-to-trough decline before liquidation & $MDD=\max_t\frac{P_{\max,t}-P_t}{P_{\max,t}}$ & Large: ``diamond hands''; Small: disciplined stop-loss & \citet{barberis2001prospect} \\
\bottomrule
\end{tabular}
\caption{Investor stylized facts: position sizing and risk control.}
\label{tab:checklist_risk}
\end{table*}

\paragraph{Layer 3 --- Market Timing and Execution Style}
Momentum versus contrarian behavior and order aggressiveness are well-studied dimensions of investor heterogeneity \citep{jegadeesh1993returns, de1985does, griffiths2000costs}.

The corresponding features and cluster interpretations are listed in Table~\ref{tab:checklist_timing}.

\begin{table*}[tp]
\setlength{\tabcolsep}{2.5pt}
\renewcommand{\arraystretch}{1.15}
\captionsetup{font=small,justification=raggedright,singlelinecheck=true}
\centering
\footnotesize
\begin{tabular}{@{}>{\raggedright\arraybackslash}p{0.13\linewidth}>{\raggedright\arraybackslash}p{0.20\linewidth}>{\raggedright\arraybackslash}p{0.27\linewidth}>{\raggedright\arraybackslash}p{0.23\linewidth}>{\raggedright\arraybackslash}p{0.12\linewidth}@{}}
\toprule
\textbf{Feature} & \textbf{Definition} & \textbf{Statistic} & \textbf{Cluster Interpretation} & \textbf{Reference} \\
\midrule
\textbf{Momentum bias} & Correlation between purchase price and recent return & $\rho_{\text{mom}}=\mathrm{Corr}(P_{\text{buy}}, R_{\text{pastN}})$ & Positive: trend follower; Negative: contrarian / reversal & \citet{jegadeesh1993returns, grinblatt2000investment} \\
\textbf{Order aggressiveness} & Fraction of volume executed via marketable orders & $AGG = V_{\text{active}}/V_{\text{total}}$ & High: news-driven / urgent; Low: limit-order / patient & \citet{griffiths2000costs} \\
\textbf{Left-/right-side trading} & Purchase position relative to local price extremum & $LR = \frac{P_{\text{buy}}-P_{\min,\text{past}}}{P_{\min,\text{future}}-P_{\min,\text{past}}}$ & Left: anticipatory / contrarian; Right: breakout / confirming & \citet{de1985does, jegadeesh1993returns} \\
\bottomrule
\end{tabular}
\caption{Investor stylized facts: market timing and execution style.}
\label{tab:checklist_timing}
\end{table*}

\paragraph{Layer 4 --- Asset Preference}

Systematic tilts toward size and volatility are robust cross-sectional patterns in investor portfolios \citep{fama1993common, ang2006cross}.

The corresponding features and cluster interpretations are listed in Table~\ref{tab:checklist_preference}.

\begin{table*}[tp]
\setlength{\tabcolsep}{2.5pt}
\renewcommand{\arraystretch}{1.15}
\captionsetup{font=small,justification=raggedright,singlelinecheck=true}
\centering
\footnotesize
\begin{tabular}{@{}>{\raggedright\arraybackslash}p{0.13\linewidth}>{\raggedright\arraybackslash}p{0.20\linewidth}>{\raggedright\arraybackslash}p{0.27\linewidth}>{\raggedright\arraybackslash}p{0.23\linewidth}>{\raggedright\arraybackslash}p{0.12\linewidth}@{}}
\toprule
\textbf{Feature} & \textbf{Definition} & \textbf{Statistic} & \textbf{Cluster Interpretation} & \textbf{Reference} \\
\midrule
\textbf{Market-cap bias} & Weighted-average log market cap of holdings & $\overline{MC}=\sum_i w_i\log MC_i$ & Large: blue-chip preference; Small: micro-/small-cap preference & \citet{fama1993common} \\
\textbf{Volatility preference} & Weighted-average historical volatility of holdings & $\bar\sigma = \sum_i w_i\sigma_i$ & High: speculative / high-beta; Low: defensive / low-vol & \citet{ang2006cross} \\
\bottomrule
\end{tabular}
\caption{Investor stylized facts: asset preference.}
\label{tab:checklist_preference}
\end{table*}

% Continue the final prose subsection without a forced column break.
\paragraph{Layer 5 --- Behavioral Finance Biases}

The disposition effect and herding are among the most pervasive behavioral biases documented in the investor-behavior literature \citep{shefrin1985disposition, odean1998investors, lakonishok1992impact, sias2004institutional}.

The corresponding features and cluster interpretations are listed in Table~\ref{tab:checklist_biases}.

\begin{table*}[tp]
\setlength{\tabcolsep}{2.5pt}
\renewcommand{\arraystretch}{1.15}
\captionsetup{font=small,justification=raggedright,singlelinecheck=true}
\centering
\footnotesize
\begin{tabular}{@{}>{\raggedright\arraybackslash}p{0.13\linewidth}>{\raggedright\arraybackslash}p{0.20\linewidth}>{\raggedright\arraybackslash}p{0.27\linewidth}>{\raggedright\arraybackslash}p{0.23\linewidth}>{\raggedright\arraybackslash}p{0.12\linewidth}@{}}
\toprule
\textbf{Feature} & \textbf{Definition} & \textbf{Statistic} & \textbf{Cluster Interpretation} & \textbf{Reference} \\
\midrule
\textbf{Disposition effect} & Tendency to sell winners too early and hold losers too long & $DE = \frac{RG}{RG+PG} - \frac{RL}{RL+PL}$ & Positive: typical retail bias; Zero/negative: rational/institutional & \citet{shefrin1985disposition, odean1998investors} \\
\textbf{Herding indicator} & Correlation of individual returns with the market & $\beta_h = \mathrm{Corr}(R_p, R_m)$ & High: trend-chaser / herder; Low: independent / contrarian & \citet{lakonishok1992impact, sias2004institutional} \\
\bottomrule
\end{tabular}
\caption{Investor stylized facts: behavioral finance biases.}
\label{tab:checklist_biases}
\end{table*}

% ============================================================
% =============================================================
%  Appendix: Scalability supplement
% =============================================================
% Flush the checklist tables before starting the scalability supplement.
\clearpage
\setcounter{secnumdepth}{3}

\subsection{Controllability Experiment Protocol}
\label{app:ctrl_setup}

Table~\ref{tab:ctrl_setup} lists the isolated-decision settings.

\begin{table}[!htbp]
  \centering
  \footnotesize
  \setlength{\tabcolsep}{4pt}
  \renewcommand{\arraystretch}{1.08}
  \begin{tabular}{@{}p{0.35\linewidth}p{0.60\linewidth}@{}}
    \toprule
    \textbf{Item} & \textbf{Setting} \\
    \midrule
    Date        & 2025-06-20$^{\dagger}$ \\
    Stock       & 300308.SZ$^{\dagger}$ \\
    Return $r_t$ & $+0.9353\%^{\dagger}$ \\
    Initial wealth $W_0$ & RMB~1,000,000 \\
    Initial allocation & 50\% stock / 50\% cash$^{\dagger}$ \\
    $\gamma$ sweep & $\{0.04,\;0.06,\;0.08,\;0.10,\;0.12\}$ \\
    Repeats per cell & 10 \\
    Total decisions & $3 \times 3 \times 5 \times 10 = 450$ \\
    Base model & Qwen3-8B \\
    Temperature & 0.6 \\
    \texttt{top\_p} / \texttt{top\_k} & 0.95 / 20 \\
    Seed & 42$^{\dagger}$ \\
    \midrule
    SUM $\gamma$ mapping & $R_{\mathrm{adj}}=\hat{R}/\gamma$ \\
    SA/TM $\gamma$ mapping & Natural-language risk profile; all other fields fixed \\
    \bottomrule
  \end{tabular}
  \caption{Controllability experiment protocol (E2).
  $^{\dagger}$Earlier-draft settings not retained in the archived
  main-sweep manifest; these require confirmation. Model and decoding
  settings are author-confirmed (Appendix~\ref{app:model_provenance}).}
  \label{tab:ctrl_setup}
\end{table}

% Use ordinary two-column text with full-width floats, without forcing a page break.
\subsection{Shared Simulation Configuration}
\label{app:sim_config}

The shared configuration is listed in Table~\ref{tab:simulation_configuration}.

\begin{table*}[t]
  \centering
  \footnotesize
  \setlength{\tabcolsep}{4pt}
  \renewcommand{\arraystretch}{1.05}
  \begin{tabular}{@{}p{0.34\linewidth}p{0.58\linewidth}@{}}
    \toprule
    \textbf{Parameter} & \textbf{Value} \\
    \midrule
    \multicolumn{2}{@{}l}{\textit{Market structure}} \\
    Stock universe & H1 E1: MSCI China A50 (50 stocks); six-day E1/E3: five-stock A-share pool \\
    Matching mechanism & Continuous double auction LOB, price--time priority \\
    Sample period & H1 E1: January 2--June 30, 2025 (117 trading days); six-day E1/E3: June 3--10, 2025 \\
    Trading frequency & Daily decision steps; cache and real-price feedback conditions vary by run \\
    \midrule
    \multicolumn{2}{@{}l}{\textit{Agent population}} \\
    Retail agents & H1 E1: SUM/RB 1000, StockAgent/TwinMarket 500, separate SUM 100 cached replay; six-day E1/E3: $N\in\{50,100,500\}$, plus SUM at 1000 \\
    Retail styles & Value / Momentum / Noise = 3:3:4 \\
    Institutional agents & One in June E3; H1 information and cache conditions in Appendix~\ref{app:h1_market_comparison} \\
    Initial cash & Retail: $\mathcal{U}[C_{\min}^{r},C_{\max}^{r}]$; institution: $\mathcal{U}[C_{\min}^{i},C_{\max}^{i}]$ \\
    \midrule
    \multicolumn{2}{@{}l}{\textit{Model and baselines}} \\
    SUM retail LLM role & Cohort-level semantic perception; utility-based retail decisions \\
    TM/SA role & Per-agent prompt-based decision baselines \\
    RB role & Non-LLM rule-feature fidelity baseline \\
    Backbone & Qwen3-8B \\
    Temperature & 0.6 \\
    \texttt{top\_p} / \texttt{top\_k} & 0.95 / 20 \\
    \bottomrule
  \end{tabular}
  \captionsetup{font=footnotesize, skip=4pt}
  \caption{Shared simulation configuration. E2 uses a single-step isolated decision protocol.}
  \label{tab:simulation_configuration}
\end{table*}
Model and decoding settings are author-confirmed
(Appendix~\ref{app:model_provenance}). Archived manifests retain anonymized
identifiers and do not independently verify this configuration.
The April 21-day scaling experiment and June 20-day date check are separate from the six-day E3 comparison reported here (Appendix~\ref{app:scaling_data_provenance}).
\subsection{Initialization of Agents, Assets, and Information}
\label{sec:initialization}
\label{app:simulation_details}

This section formalizes the initialization procedures for heterogeneous agents, market assets, and the exogenous news dataset. The design aims to ensure (i) statistically grounded heterogeneity, (ii) economically interpretable initial states, and (iii) a consistent information-generation process that drives market dynamics.

\subsubsection{Retail Agent Initialization}

Let $\mathcal{A}_R = \{1, \dots, N_R\}$ denote the set of retail agents. Each agent $i \in \mathcal{A}_R$ is initialized with a portfolio state, behavioral parameters, and latent heterogeneity variables.

\paragraph{Wealth Distribution.}
To reflect empirically observed long-tailed wealth distributions, initial wealth $W_i^{(0)}$ is sampled from a truncated Pareto distribution:
\begin{equation}
\begin{gathered}
W_i^{(0)}=\min\left(\frac{W_{\min}}{(1-U_i)^{1/\alpha}},\,W_{\max}\right),\\
U_i\sim\mathcal{U}(0,1),
\end{gathered}
\end{equation}
where $W_{\min}$ and $W_{\max}$ denote lower and upper truncation bounds, and $\alpha > 1$ controls tail heaviness.

\paragraph{Asset Allocation.}
Each agent allocates an equity share $\rho_i$ and samples $K_i$ assets from $\mathcal{S}$. Portfolio weights follow:
\begin{equation}
\mathbf{w}_i \sim \text{Dir}(\mathbf{1}_{K_i}), \quad \sum_{k=1}^{K_i} w_{i,k} = 1.
\end{equation}
The capital allocated to asset $s_k$ is then
\begin{equation}
M_{i,k} = W_i^{(0)} \cdot \rho_i \cdot w_{i,k}.
\end{equation}

\paragraph{Initial Positions and Cost Basis.}
Given initial market prices $P_{s_k}^{(0)}$, the position size and cost basis are defined as:
\begin{align}
Q_{i,k}^{(0)} &= \left\lfloor \frac{M_{i,k}}{P_{s_k}^{(0)}} \right\rfloor, \\
C_{i,k}^{(0)} &= P_{s_k}^{(0)} \cdot (1 + \epsilon_{i,k}), 
\quad \epsilon_{i,k} \sim \mathcal{N}(0, \sigma_{\text{price}}^2).
\end{align}
This construction induces heterogeneous psychological anchors consistent with behavioral finance evidence.

\paragraph{Micro-level Heterogeneity.}
Each agent is further parameterized by:
\begin{align}
\gamma_i &\sim \mathcal{U}[0.04,0.12] \quad \text{(risk damping)}, \\
\sigma_i &\sim \mathcal{U}[0.005,0.030] \quad \text{(perception noise)}.
\end{align}
These parameters govern decision sensitivity and informational precision in downstream dynamics.

\subsubsection{Institutional Agent Initialization}

Let $\mathcal{A}_I$ denote the set of institutional agents. Rather than independent sampling, institutional agents are initialized as aggregate and strategic counterparts to retail positions.

\paragraph{Aggregate Portfolio Construction.}
The initial institutional portfolio is constructed by aggregating retail holdings:
\begin{equation}
Q_{s}^{I,(0)} = \sum_{i \in \mathcal{A}_R} Q_{i,s}^{(0)}.
\end{equation}

\paragraph{Cost Basis Formation.}
Institutional cost bases reflect strategic accumulation under favorable conditions:
\begin{equation}
C_s^{I,(0)}=\min_{k\in\{5,10,20,30\}}\text{MA}_k(s),
\end{equation}
where $\text{MA}_k(s)$ denotes the $k$-period moving average price of asset $s$. This models patient capital deployment consistent with institutional trading behavior.

\subsubsection{Asset Initialization}

Let $\mathcal{S} = \{1, \dots, M\}$ denote the asset universe. Each asset $s \in \mathcal{S}$ is initialized with:

\begin{itemize}
    \item Initial price $P_s^{(0)}$ derived from historical market data,
    \item Fundamental value proxy $V_s^{(0)}$,
    \item Historical return series used to compute volatility:
    \begin{equation}
    \sigma_s^2 = \text{Var}(r_{s,t}) \quad \text{over a rolling window}.
    \end{equation}
\end{itemize}

These quantities provide the sufficient statistics required by utility-based decision modules.

\subsubsection{News Dataset Construction}

The information process is modeled as a structured, time-indexed dataset that drives agent perception.

\paragraph{Data Schema.}
Each news item is represented as a tuple:
\begin{equation}
\mathcal{N}_t = \{ (\text{scope}, \text{timestamp}, \text{content}) \},
\end{equation}
where:
\begin{itemize}
    \item $\text{scope} \in \{\text{macro}\} \cup \mathcal{S}$ specifies whether the news is market-wide or asset-specific,
    \item $\text{timestamp}$ is defined at minute-level granularity,
    \item $\text{content}$ is unstructured textual information.
\end{itemize}

\paragraph{Sampling and Filtering Mechanism.}
At each time step $t$, agents observe a filtered subset of the global news stream $\mathcal{S}_t$:
\begin{equation}
\mathcal{O}_{i,t} = \text{TopK} \left( \mathcal{S}_t, \; \text{Sim}(\text{Profile}_i, \text{Content}) \right),
\end{equation}
where $\text{Sim}(\cdot, \cdot)$ denotes semantic similarity and $\text{TopK}$ selects the most relevant items. This induces heterogeneous information exposure aligned with agent profiles.

\paragraph{Distributional Design.}
The dataset integrates:
\begin{itemize}
    \item \textbf{Macro-level signals}: policy changes, interest rates, and global events,
    \item \textbf{Firm-level signals}: earnings, announcements, and sector-specific developments.
\end{itemize}
Retail news inputs follow the daily information cutoff. The archived June
scaling runs give the institutional agent next-day news in advance
(Appendix~\ref{app:shared_experimental_settings}); these runs therefore do not provide
a strictly real-time forecasting test.

\paragraph{Role in the System.}
The constructed news dataset serves as the primary exogenous driver of belief updates, enabling the perception module (System 1) to transform unstructured signals into structured expectations, which are subsequently consumed by utility-based decision processes.

% \medskip

Collectively, these initialization procedures establish a heterogeneous yet controlled environment, ensuring that both cross-sectional diversity and temporal information flow are grounded in economically interpretable mechanisms.

\subsection{System Comparison Table}
\label{app:comparison_table}

Table~\ref{tab:comparison} summarizes the retail-agent implementations
used in this paper. The StockAgent and TwinMarket adaptations are detailed
in Appendix~\ref{app:baselines}.

\begin{table*}[t]
\centering
\footnotesize
\setlength{\tabcolsep}{5pt}
\renewcommand{\arraystretch}{1.05}
\begin{tabular}{p{0.13\textwidth}p{0.20\textwidth}p{0.17\textwidth}p{0.12\textwidth}p{0.18\textwidth}}
\toprule
\textbf{System} &
\textbf{Retail Decision Design} &
\textbf{Evaluated $N$} &
\textbf{Retail LLM Unit} &
\textbf{Behavioral Interface} \\
\midrule
StockAgent &
Direct-action prompting &
50, 100, 500 &
Per-agent &
Risk profile and direct order generation \\
TwinMarket &
Belief--desire--intention prompting &
50, 100, 500 &
Per-agent &
Risk profile, belief, intention, and action \\
Rule-based &
Explicit trading rules &
1000 &
None &
Structured market features and trading rules \\
\textbf{SUM (ours)} &
Shared perception and utility-based decisions &
50, 100, 500, 1000 &
Per-cohort &
Shared semantic signals and individual utility parameters \\
\bottomrule
\end{tabular}
\captionsetup{justification=raggedright, singlelinecheck=false, font=footnotesize}
\caption{Comparison of the evaluated retail-agent pipelines.
Agent counts are pooled across the H1 and six-day experiments;
the rule-based baseline is evaluated only in H1.
The table describes our implementations, not the full architectures or
reported scales of the original systems.}
\label{tab:comparison}
\end{table*}

% Appendix C: original A.2 and A.3.
% Original A: Ablation and Supplementary Experiments
\section{} % TODO(author): fill in title.
\label{app:supplementary_market_experiments}

% Original A.2: Heterogeneity Ablation Experiments
\subsection{} % TODO(author): fill in title.
\label{sec:ablation}

\paragraph{Settings.}
We compare Full SUM with three heterogeneity interventions, using
1,000 retail agents, five stocks, and seed 42. The April window has
21 trading days and the June window has 20. The stock baskets differ,
so this is not a date-only controlled comparison. Macro preserves
institutional capital but replaces the institutional decision structure.
Meso unifies retail agents under the Noise signal--behavior channel;
it does not remove the Value or Momentum utility in isolation.
Micro sets the initial risk parameter to $\gamma=0.08$ and the perception
noise parameter to $\sigma=0.0175$, their distribution means.
This initialization does not imply that the effective risk parameter
remains fixed throughout the run.

\begin{table}[t]
  \centering
  \small
  \setlength{\tabcolsep}{4pt}
  \renewcommand{\arraystretch}{1.05}
  \begin{tabular}{@{}lrr@{}}
    \toprule
    Condition
      & \shortstack{April 2025\\21 days\\EW-RMSE $\downarrow$}
      & \shortstack{June 2025\\20 days\\Return W1 (pp) $\downarrow$} \\
    \midrule
    Full SUM & \textbf{55.02} & \textbf{0.893} \\
    Macro    & 58.35         & 0.918          \\
    Meso     & 57.69         & 1.163          \\
    Micro    & 56.86         & 0.907          \\
    \bottomrule
  \end{tabular}
  \caption{Selected ablation results from two separate five-stock windows
  ($N=1000$, seed 42). April reports equal-weight index RMSE in base-1000
  index points; June reports the W1 distance between simple daily-return
  distributions in percentage points. The window--metric pairs were
  selected after inspecting the results. Bold marks the lowest value
  within each column; the columns are not the same metric across dates.}
  \label{tab:ablation_april}
  \label{tab:ablation_selected}
\end{table}

\paragraph{Results.}
Table~\ref{tab:ablation_selected} reports the April metrics from the
exported comparison table and the June return distances reconstructed
from the archived stock-level real and simulated prices. Full SUM has
the lowest April index RMSE and June return W1 among these configurations.
The interventions therefore weaken the corresponding outcome in each
reported window. These selected, single-seed results do not establish
that every intervention worsens every metric, or that one heterogeneity
level alone causes a particular market property.

% Original A.3: U.S. Equity Transfer Experiment
\subsection{} % TODO(author): fill in title.
\label{app:us_transfer}

We conduct a preliminary transfer experiment on the U.S. Magnificent 7
over June 2025. Both the U.S. and A-share SUM runs use
$N=500$ agents and $M=3$ cohorts; this does not establish that
all other settings are identical. Table~\ref{tab:us_transfer}
reports the selected archived metrics for this comparison,
which is separate from the H1 market comparison and six-day scaling runs.

\begin{table}[tbp]
  \centering
  \small
  \setlength{\tabcolsep}{4pt}
  \renewcommand{\arraystretch}{1.10}
  \begin{tabular*}{\linewidth}{@{\extracolsep{\fill}}lrr@{}}
    \toprule
    Metric & A-share & \shortstack{U.S.\\Magnificent 7} \\
    \midrule
    NRMSE $\downarrow$ & \textbf{0.0296} & 0.0833 \\
    RetCorr $\uparrow$ & \textbf{0.3885} & 0.3787 \\
    \bottomrule
  \end{tabular*}
  \captionsetup{font=10pt,skip=4pt}
  \caption{Selected preliminary cross-market results for SUM
  ($N=500$, $M=3$, June 2025). Arrows indicate metric directions;
  bold marks the better value in each row. Values are archived reports,
  not estimates recomputed from daily U.S. price paths.}
  \label{tab:us_transfer}
\end{table}

The U.S. run has a return correlation close to, but lower than, the
A-share result (0.3787 vs.\ 0.3885). Its normalized price-path error is
larger: NRMSE increases from 0.0296 to 0.0833. This is a preliminary
cross-market check, not evidence of an overall improvement after transfer
or a comparison against other methods in the U.S. market.
The archive contains these reported metrics but no matching U.S. daily
price series. We therefore retain the reported normalization rather than
claiming an independent recomputation or importing settings from the H1
and six-day experiments.

% Source checked against COLM 2026 author replies HktklnUOSK and XIgOMKenCP:
% https://openreview.net/forum?id=AOVW01GBKJ&noteId=XIgOMKenCP
% Both replies contain the same four-row table. Keep this provenance out
% of the compiled anonymous manuscript.
% TODO before submission: verify the exact stock list, trading dates,
% LLM/decoding settings, seeds/repeats, and normalization formulas against
% the original U.S. run. These details and daily paths are not present in
% the archived U.S. result folder; do not import settings from other runs.

\section{Implementation and Experimental Details}
\label{app:implementation_details}

\subsection{Prompt Templates}
\label{app:prompt_templates}
% TODO: double check prompt templete before submit.

This section presents the prompt templates in English.
Placeholders are filled with the information available at each decision step.
SUM retail prompts are provided in Appendix~\ref{app:sum_retail_prompts}.
Institutional and baseline prompts are grouped separately below. The rule-based baseline does not use an LLM prompt.

\subsubsection{Institutional Agent Prompts}
\label{app:institution_prompts}

The institutional agent uses a hierarchical LLM workflow: a fundamental analyst,
a quantitative strategist, a sentiment-and-flow analyst, and a CIO aggregator.
The prompt fields below match the implemented code, but the prose is translated
from Chinese to English.

\paragraph{Fundamental analyst: quarterly baseline.}
\begin{Verbatim}[breaklines=true,breakanywhere=true,fontsize=\footnotesize]
You are a fundamental analyst at an asset-management firm. Build an intrinsic
value baseline and earnings-expectation framework for this company, to be used
by daily incremental updates over the next several weeks.

[Company profile]
{firm_name} ({industry})
Main business: {main_business}

[Latest quarterly financial data]
Income statement: revenue, net income, gross margin, net margin, EPS
Balance sheet: total assets, total liabilities, debt/assets, current ratio, BPS
Cash flow: operating cash flow, free cash flow
Composite metrics: ROE, ROA

[Macro environment]
10Y government bond yield; 1Y LPR

[Tasks]
1. Estimate a reasonable P/B range (low, midpoint, high) and explain the main
   drivers.
2. List 3-5 key factors that may change future earnings expectations.

Return valid JSON only:
{
  "fundamental_quality": float,
  "fair_pb_mid": float,
  "fair_pb_range": [float, float],
  "key_expectation_factors": [string],
  "risk_flag": float,
  "context_summary": string
}
\end{Verbatim}

\paragraph{Fundamental analyst: daily expectation update.}
\begin{Verbatim}[breaklines=true,breakanywhere=true,fontsize=\footnotesize]
You are tracking daily expectation changes for {firm_name}.

[Quarterly baseline]
context_summary = {context_summary}
key_expectation_factors = {key_expectation_factors}

[Current valuation]
current P/B; fair P/B midpoint; valuation deviation

[Today's new information, including early-access news]
{news_inst_text}

[Tasks]
1. Decide whether today's information raises or lowers expected intrinsic value.
2. Assess whether the change is evidence-based or speculative.
3. Decide whether expectation improvement can still move price even if the
   stock is already expensive.

Return valid JSON only:
{
  "expectation_delta": float,
  "delta_confidence": float,
  "catalyst_type": "policy" | "industry" | "company_event" |
                   "macro" | "competitor" | "none"
}
\end{Verbatim}

\paragraph{Quantitative strategist prompt.}
\begin{Verbatim}[breaklines=true,breakanywhere=true,fontsize=\footnotesize]
You are a quantitative strategist. Judge short-horizon price movement from
technical and capital-flow information.

[Company]
{firm_name} ({industry})

[Technical state]
20-day close-price ratios; 20-day volume ratios; MA5/P_t; MA20/P_t;
order-book imbalance (OBI); range position

[Capital-flow data]
Margin-financing balance change; securities-lending balance change;
block-trade information if available

[Tasks]
1. Analyze trend direction and strength, including momentum, reversal,
   support, and resistance.
2. Analyze capital-flow signals.
3. Forecast the 1-5 day short-term price direction.

Return valid JSON only:
{
  "technical_signal": float,
  "capital_flow_signal": float,
  "volatility_forecast": float,
  "key_levels": {"support": float, "resistance": float},
  "context_summary": string
}
\end{Verbatim}

\paragraph{Sentiment and flow analyst prompt.}
\begin{Verbatim}[breaklines=true,breakanywhere=true,fontsize=\footnotesize]
You are a behavioral-finance analyst studying retail behavior and arbitrage
opportunities.

[Retail monitoring data]
number of retail holders; net buyer count today; average retail cost basis;
dispersion of cost basis; mean and variance of noise-agent sentiment;
bullish/bearish/neutral forum counts

[Public forum posts]
{posts_text}

[News, including early-access news]
{news_inst_text}

[Market dynamics]
volume ratio; OBI

[Tasks]
1. Measure whether retail sentiment is extremely bullish or bearish.
2. Evaluate whether many retail investors are in unrealized gains or losses.
3. Identify possible retail stampede or bottom-fishing behavior.
4. Decide how an institution could exploit these signals contrarily.

Return valid JSON only:
{
  "retail_sentiment_extreme": float,
  "retail_pain_signal": float,
  "contrarian_opportunity": float,
  "context_summary": string
}
\end{Verbatim}

\paragraph{CIO aggregation prompt.}
\begin{Verbatim}[breaklines=true,breakanywhere=true,fontsize=\footnotesize]
You are the chief investment officer. Combine the three analyst reports and make
the final portfolio decision.

[Target]
{firm_name}; current price P_t

[Analyst reports]
Fundamental analyst: quality, fair P/B midpoint, valuation state,
expectation_delta, confidence, catalyst type, risk flag, summary
Quantitative strategist: technical signal, capital-flow signal, volatility
forecast, support, resistance, summary
Sentiment/flow analyst: retail sentiment extreme, retail pain signal,
contrarian opportunity, summary

[Current position]
current shares; position value; stock weight; total portfolio exposure;
single-stock limit

[Decision requirements]
1. Identify the main disagreement among analysts.
2. Balance expected return and risk control.
3. Respect current holdings and portfolio constraints.
4. Be more aggressive when analysts agree; be conservative under disagreement.
5. Pay special attention to high-confidence retail-behavior arbitrage signals.

Return valid JSON only:
{
  "action": "buy" | "sell" | "hold",
  "conviction": float,
  "target_weight_delta": float,
  "urgency": float,
  "risk_override": bool,
  "reasoning": string
}
\end{Verbatim}

\subsubsection{Baseline Prompt Templates}
\label{app:prompt_baseline_templates}

\paragraph{StockAgent-style direct-action template.}
\begin{Verbatim}[breaklines=true,breakanywhere=true,fontsize=\footnotesize]
System:
You are a {style}-type retail investor in the Chinese A-share market. Strictly
follow the given persona, account constraints, and trading format.

Risk profile: {gamma_profile_text}
Judgment stability: {sigma_profile_text}
Investment style: {style_desire_text}

Act like a StockAgent-style trader: directly decide one trading action from the
current market state. Return valid JSON only. If no trade is needed, return
{"orders": [], "post_text": "..."}.

User:
[Target] {firm_name} ({stock_code})
[Market state] {market_state_text}
[News] {news_text}
[Forum posts] {posts_text}

[Price and structured indicators]
current_price; kline_ratios; P/B; BPS; range_pos; OBI; vol_ratio

[Account and trading constraints]
cash; current shares; current position value; wealth; max trade value ratio;
price tick; price limits; minimum order quantity

[Hard numeric constraints]
- If orders is non-empty, deal_amount must be positive.
- BUY amount must not exceed min(cash * 0.95, wealth * max_trade_value_ratio).
- SELL amount must not exceed min(current_shares * current_price,
  wealth * max_trade_value_ratio).
- If no trade is warranted, return an empty orders array.
- Use style-specific limit-price offset guidance.

Return valid JSON only:
{
  "orders": [{
    "stock": "{stock_code}",
    "direction": "BUY or SELL",
    "price_type": "LIMIT",
    "price_offset": float,
    "deal_amount": float,
    "agent_id": "{agent_id}",
    "client_order_id": "ORD-{sim_step}-{agent_id}-{stock_code}",
    "reasoning": string
  }],
  "post_text": string
}
\end{Verbatim}

\paragraph{TwinMarket-style BDI template.}
\begin{Verbatim}[breaklines=true,breakanywhere=true,fontsize=\footnotesize]
System:
You are a personalized A-share retail investor following the TwinMarket
decision style:
1. maintain and update Belief;
2. follow the Desire block, risk preference, and investment style;
3. form Intention and output the final action.
Your investor type must remain {style}. Return valid JSON only.

User:
[Previous belief]
{belief_prev}

[Today's observable information]
target stock; market state; news digest; forum posts; current price;
20-day price series; P/B and BPS; range position; OBI; volume ratio

[Desire]
{style_desire_text}
Risk preference: {gamma_profile_text}
Information-processing stability: {sigma_profile_text}

[Account and trading constraints]
cash; current shares; current position value; wealth; max trade value ratio;
price tick; price limits; minimum order quantity

[Hard numeric constraints]
Same positive amount, cash, inventory, and style-specific offset constraints as
the StockAgent-style template.

Return valid JSON only:
{
  "belief": "updated first-person belief, 1-3 sentences",
  "intention": "one-sentence trading intention",
  "orders": [{
    "stock": "{stock_code}",
    "direction": "BUY or SELL",
    "price_type": "LIMIT",
    "price_offset": float,
    "deal_amount": float,
    "agent_id": "{agent_id}",
    "client_order_id": "ORD-{sim_step}-{agent_id}-{stock_code}",
    "reasoning": string
  }],
  "post_text": string
}
\end{Verbatim}

\subsection{Experiment Design}
\label{app:experiment_design}

\subsubsection{Shared Environment and Information Access}
\label{app:baseline_input_alignment}
\label{app:shared_experimental_settings}

SUM and the prompt-based baselines use the A-share continuous
double-auction limit-order-book simulator, with price--time priority
and a common order interface. Their orders follow cash, inventory,
price-tick, price-limit, and minimum-lot constraints. The rule-based
baseline is a separate implementation. The stock universe, trading
dates, agent population, and initial conditions are specified for each
experiment; matching dates and stocks does not make all configurations
identical.

At a matched decision state, SUM and the prompt-based baselines receive
the same available news, forum text, market indicators, and account fields.
In closed-loop experiments, prices, accounts, and other endogenous states
can differ as each simulation follows its own trading history.
The rule-based baseline uses structured market, accounting, and forum
statistics rather than news or forum text.
Retail information access is restricted to the corresponding decision-time
window. The June scaling runs also include one institutional agent with
next-trading-day news access, an explicit simulation assumption
(Appendix~\ref{app:institution_agent}). Its decisions are replayed from the
same institutional cache for the two prompt-based baselines at each $N$.
This setting is not a strictly real-time forecasting test.
The H1 comparison also includes next-trading-day institutional news
and weekly feedback from real market prices; its run and cache
conditions are described in Appendix~\ref{app:h1_market_comparison}.

The original initialization samples
$\gamma_0 \sim \mathrm{Uniform}(0.04,0.12)$ and
$\sigma \sim \mathrm{Uniform}(0.005,0.030)$, with $\psi$ sampled by investor type.
The controllability experiment uses a fixed $\gamma$ sweep instead.
% TODO: Confirm the distributions actually used in each final run.
% TODO: Specify agent counts, type proportions, initial wealth/holdings,
%       institutional participation, decision frequency, and forum activation
%       separately for E2, E3, and E1.
% TODO: Add the model version, decoding settings, seed handling, and repetition
%       protocol. Do not claim that all experiments have identical settings.

\subsubsection{Controllability Experiment}
\label{app:controllability_design}

The controllability experiment uses a single-step decision task.
For each investor type, the market snapshot and account state are held fixed
while the risk parameter $\gamma$ is varied.
SUM uses $\gamma$ in its decision rule, while the prompt-based baselines
encode it through natural-language risk profiles.
The evaluation measures the ordering of position adjustments across risk
levels and the variation across repeated decisions.
The main sweep uses five risk levels and ten repeats per cell. Within-level
standard deviations use the sample denominator $10-1$.
The separate robustness runs vary temperature (0.0 and 0.7) and market
scenario; their settings are not assumed to apply to the main sweep.
Appendix~\ref{app:ctrl_setup} distinguishes verified fields from missing
main-sweep metadata.

% TODO: Consolidate the existing E2 protocol (Appendix H / Table 8) here.
%       The latest main-text protocol specifies stock 300308.SZ on 2025-06-20,
%       initial wealth RMB 1,000,000, a 50% stock / 50% cash allocation,
%       gamma in {0.04, 0.06, 0.08, 0.10, 0.12}, and 10 repeats per cell.
%       Confirm these against the final run before copying them into prose.
% TODO: Explain whether belief states, perception outputs, and private noise
%       are reset, fixed, or resampled between settings and repetitions.

\subsubsection{Scalability Experiment}
\label{app:scaling_data_provenance}

The experiment ledger separates three scaling records.
The main April experiment covers April 1--30, 2025 (21 trading days,
five stocks, seed 42) and compares SUM with a Pure LLM baseline.
An independent June date check covers 20 trading days with a different
five-stock pool. These are separate experiments, not points on one curve.

Figures~\ref{fig:pilot_scalability} and~\ref{fig:resource_cost} and
Table~\ref{tab:main_market_results}(b) use the archived six-day StockAgent and
TwinMarket comparison from June 3, 4, 5, 6, 9, and 10, 2025.
These are the first six trading days recorded in the selected runs; we use
the same dates for token totals and index metrics, excluding initialization
from RMSE. They are not full-month June results or replacements for April.
The baseline runs end on June 10; SUM uses the first six daily records of
the uncached June runs. The stocks are 002594.SZ, 600519.SH, 300059.SZ,
300502.SZ, and 300308.SZ. At $N=50$, the baseline measurements use the
original runs, not later cached replays. The archive does not document a
further reason for this short window.
SUM is evaluated at $N=50,100,500,1000$ and the prompt-based methods at
$N=50,100,500$, with one run per setting and seed 42.

SUM's recorded tokens include retail cohort perception and institutional
inference. Baseline tokens include retail input and output tokens but exclude
generation of their replayed institutional cache. SUM uses 90 retail
cohort-perception calls at each scale. At $N=500$, each baseline makes
10,710 retail decision calls: the ratio of 119 compares these retail
interfaces, not total system calls. Retries, hardware, concurrency, and
end-to-end cache-generation costs require separate accounting.
At $N=500$, the recorded totals are 533,924 tokens for SUM, 36,858,039
for StockAgent, and 38,364,657 for TwinMarket.

% TODO: Verify the run manifests and distinguish SUM, SA, and TM explicitly.
%       The pasted draft refers to a "Pure LLM" summary, which must be matched
%       to the final baseline implementations before it is used as evidence.
% TODO: Specify the exact six trading dates, activation/skip rules, invocation
%       frequency, cold/warm cache policy, and which calls/tokens are counted.
% TODO: State the hardware, concurrency, timing boundary, and retry policy.

\paragraph{Resource projections.}
\label{app:projection}
Measured runs and projected resource costs are reported separately.
No simulation-fidelity score is assigned to an unexecuted population size.

% TODO: Confirm the projection estimator against the final resource logs.
%       The pasted draft fits y_N = aN + b at N in {50,100,500} and reports
%       329.16M tokens, 7.83 hours, and 71,359 calls for "Pure LLM" at N=1000.
%       These values differ from the latest main-text SA/TM projections.
%       They are intentionally not repeated as verified settings here.

\subsubsection{Market Fidelity Experiment}
\label{app:fidelity}
\label{app:fidelity_data_provenance}

Table~\ref{tab:main_market_results}(a) reports selected H1 metrics from
50 MSCI China A50 stocks and 117 trading days. The archived daily
index files are available for all compared runs. SUM $N=1000$ and
Rule-based $N=1000$ use their full-year runs restricted to H1;
StockAgent and TwinMarket use H1 runs with $N=500$ each. The SUM
$N=100$ W1 result is a separate H1 cached replay, not the $N=1000$
configuration. All runs are scored against the StockAgent $N=500$
real equal-weight index. Appendix~\ref{app:h1_market_comparison}
defines the metrics, cache conditions, and selected-result scope.

Figure~\ref{fig:emergent_fidelity} is a separate January--June SUM
diagnostic from the archived figure series. Its curve and metrics
must not be substituted for the H1 comparison above. The full-year
investor analysis in Appendix~\ref{sec:fidelity_appendix} also uses
separate records. Heterogeneity interventions are reported in
Appendix~\ref{sec:ablation}, and the U.S. transfer check in
Appendix~\ref{app:us_transfer}.

\subsubsection{Evaluation Metrics}
\label{app:metric_definitions}

The simulated and empirical equal-weighted price indices are normalized
to 1000 at initialization. For the June scaling comparison, RMSE uses
the six daily closes and excludes the initialization row.
The H1 panel uses daily simple returns, population standard deviations,
and a common real-index target as specified in
Appendix~\ref{app:h1_market_comparison}. Its SD error and return W1
are in percentage points. The April ablation reports index RMSE,
and the June ablation reports simple-return W1; these metrics and
windows are not interchangeable.
Market-level metrics include cumulative return over the evaluation window,
root mean squared index error, mean absolute percentage error,
daily return correlation, and direction accuracy.
Volatility clustering is measured by the autocorrelation of absolute daily
returns, $|r_t|$, at the reported lags.
The volume--volatility diagnostic measures the Pearson correlation between
simulated daily trading volume and absolute daily returns.

Investor-level metrics include the Gini coefficient, wealth shares,
counterparty shares, and investment returns.
The Gini coefficient is computed from the cross-sectional wealth distribution
at each day. A style's wealth share is its fraction of total simulated wealth.
Counterparty shares are computed from executed transaction values and
normalized across counterparties for each focal investor type.
The investment return of retail agent $i$ is
$(W_{i,T}-W_{i,0})/W_{i,0}$.

% TODO: Keep exact E2 metric definitions, resource accounting, and any
%       experiment-specific metric conventions consistent with the main text.

% ==============================================================================
\subsubsection{Detailed Settings for Main Experiments}
\label{app:main_experiment_setups}

\paragraph{E1: Simulation quality.}
Table~\ref{tab:main_market_results}(a) reports the 50-stock MSCI China A50
comparison over H1 2025 (January 2--June 30, 117 A-share trading days;
seed 42). SUM and the rule-based baseline use $N=1000$ retail agents;
StockAgent and TwinMarket each use $N=500$. The reported SUM W1
uses a separate $N=100$ cached replay. All systems are compared with
the StockAgent $N=500$ real equal-weight index, with selected metrics
and run conditions disclosed in Appendix~\ref{app:h1_market_comparison}.
Panel (b) reports a separate matched-agent comparison over five stocks and
six trading days (June 3--10, 2025; seed 42), using the same runs as E3.
SUM runs at $N\in\{50,100,500,1000\}$; StockAgent and TwinMarket run at
$N\in\{50,100,500\}$. RMSE uses six daily closes and excludes initialization.
Figure~\ref{fig:emergent_fidelity} reports a separate archived H1 2025
diagnostic (January--June; 117 trading days), not the same
common-target comparison as panel (a).
The investor-level results use separate full-year 2025 records.
Appendix~\ref{sec:ablation} consolidates the April and June heterogeneity
ablations. Appendix~\ref{app:us_transfer} reports the separate June
U.S. equity transfer check.

\paragraph{E2: Controllability.}
We use a fixed single-asset decision task with
$W_0{=}\text{RMB }1\text{M}$ and the model configuration in
Appendix~\ref{app:model_provenance}.
We test $\gamma\in\{0.04,0.06,0.08,0.10,0.12\}$ for three investor types
and three systems, with ten repeats per cell (450 decisions).
SUM maps $\gamma$ as $R_{\mathrm{adj}}{=}\hat{R}/\gamma$;
StockAgent and TwinMarket encode it as a natural-language risk profile.
Appendix~\ref{app:ctrl_setup} gives the protocol and identifies setup
details not retained in the archived main-sweep manifest.

\paragraph{E3: Scalability.}
We use the same five-stock A-share pool and six trading days
(June 3--10, 2025; seed 42) as the matched-agent E1 comparison and
Section~\ref{sec:pilot_scalability}.
These six-day measurements are separate from the April 21-day SUM--Pure LLM
comparison and the June full-month date check
(Appendix~\ref{app:scaling_data_provenance}).
The model configuration is given in Appendix~\ref{app:model_provenance}.
Figure~\ref{fig:resource_cost} shows measured runs only; $N=1000$ baseline
cost projections are described in Appendix~\ref{app:projection_method}.
SUM's recorded tokens include cohort perception and institutional inference.
Baseline tokens cover retail inference but exclude generation of the replayed
institutional cache. The 90 SUM calls count cohort perception only and
exclude institutional calls.

% Part 2: originally content/fidelity_appendix.tex
% ==============================================================================
\section{Additional Fidelity Diagnostics}
\label{sec:fidelity_appendix}

\subsection{Investor-Level Style Matching}

Table~\ref{tab:investor_fidelity} reports investor-level diagnostics across all
four systems.
The runs cover different horizons: SUM/RB cover the full year, StockAgent
covers January 2--April 22, and TwinMarket covers Q1.
We describe differences between styles within each run, not performance
rankings across systems.

\begin{table}[htbp]
  \centering
  \scriptsize
  \setlength{\tabcolsep}{3.5pt}
  \renewcommand{\arraystretch}{1.08}
  \begin{tabular}{@{}lrrrr@{}}
    \toprule
    \textbf{Style}
      & \textbf{SUM} & \textbf{SA} & \textbf{TM} & \textbf{RB} \\
    \midrule
    Noise        & 102.3\% & 7.6\% & 7.8\% & 42.8\% \\
    Momentum     & 95.1\%  & 7.1\% & 8.6\% & 58.3\% \\
    Value        & 76.9\%  & 5.9\% & 6.4\% & 29.7\% \\
    Institution  & 17.9\% & 4.5\% & 4.2\% & 11.1\% \\
    \bottomrule
  \end{tabular}
  \captionsetup{font=footnotesize, skip=4pt}
  \caption{Aggregate wealth returns by investor style.
  SUM and Rule-based (RB) cover the full year (243 trading days);
  StockAgent (SA) covers January 2--April 22 (72 trading days);
  TwinMarket (TM) covers Q1 (57 trading days), all in 2025.
  The different horizons prevent direct comparisons of return magnitudes
  across systems.}
  \label{tab:investor_fidelity}
\end{table}

Noise agents have higher aggregate wealth returns than momentum and value
agents in the SUM and StockAgent runs. Momentum agents have the highest
retail-style returns in the TwinMarket and Rule-based runs.
These are observed within-run differences, not a universal expected ranking
or a validation criterion. SUM's institutional agent returns 17.9\% over
the full year, within 1.0 percentage point of the 16.9\% return of that
run's real equal-weighted component basket.

\subsection{Micro-Control Validity}

The micro-control audit measures how $\gamma_i$ and $\sigma_i$ are associated
with returns in a separate full-year 2025 run with 1,000 retail agents.
These closed-loop associations complement the single-step test in E2 but do
not establish causal effects on investment returns.

\begin{figure}[htbp]
  \centering
  \includegraphics[width=\linewidth]{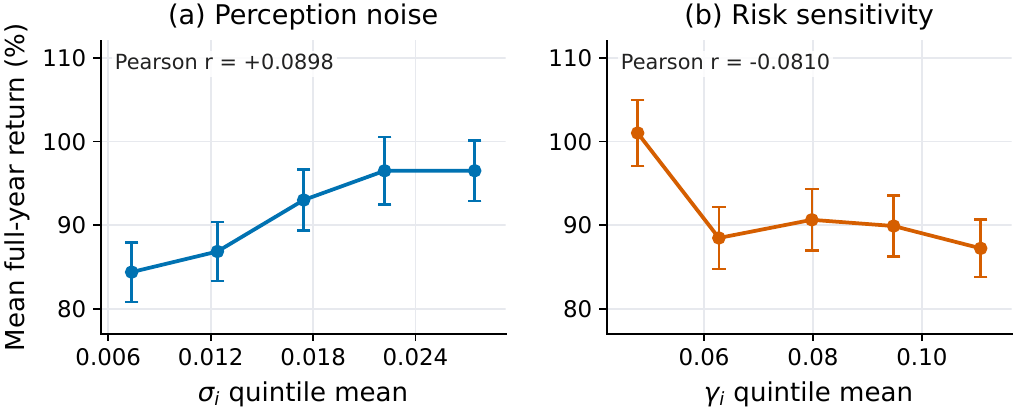}
  \captionsetup{font=footnotesize}
  \caption{Full-year 2025 micro-control diagnostics for 1,000 retail agents.
  Points show mean returns within five equal-sized parameter groups
  (200 agents each); error bars show standard errors across agents within
  each group, not uncertainty across simulation seeds.
  Pearson correlations use all 1,000 individual agents:
  $r{=}+0.0898$ for $\sigma_i$ and $r{=}-0.0810$ for $\gamma_i$.}
  \Description{Two quintile-mean plots for full-year returns. Returns generally
  rise with perception noise and fall with risk sensitivity, with a reversal
  between the second and third risk-sensitivity groups.}
  \label{fig:micro_controls}
\end{figure}

Risk sensitivity has a weak negative association with full-year returns
(Pearson $r{=}-0.0810$ overall and $r{=}-0.326$ within the momentum cohort).
Mean returns from the lowest to the highest $\gamma_i$ quintile are
101.00\%, 88.45\%, 90.63\%, 89.88\%, and 87.21\%.
The overall trend is downward, but the increase from the second to the third
group rules out strict monotonicity. Perception noise has a weak positive
association with returns ($r{=}+0.0898$). These descriptive results do not
show that either parameter alone determines investment performance.

% ==============================================================================
% Part 4: originally content/institution_agent_appendix.tex
% ==============================================================================
\section{Institution Agent Structure}
\label{app:institution_agent}

\subsection{Overview}

The institution agent is modeled as an informed, rational decision-maker with a hierarchical internal structure and access to enriched information. Unlike retail agents, whose decisions are driven by a single LLM-utility pipeline, the institution agent employs a multi-stage aggregation mechanism that integrates heterogeneous signals under explicit portfolio and execution constraints.

At each discrete time step $t$, the institution agent produces trading decisions over a set of assets $\mathcal{J} = \{1, \dots, M\}$.

\subsection{Information Structure}

For each asset $j \in \mathcal{J}$ at time $t$, the institution agent observes an information set:
\begin{equation}
\mathcal{I}_{j,t}^{\mathrm{inst}} = \left\{
\mathcal{N}_{t}, \mathcal{N}_{t+1}, \mathcal{P}_{t-1}^{\mathrm{inst}}, \mathcal{S}_{t}^{\mathrm{ret}}, \mathcal{F}_{j,t}, \mathcal{M}_{j,t}
\right\},
\end{equation}
where:
\begin{itemize}
    \item $\mathcal{N}_{t}$ and $\mathcal{N}_{t+1}$ denote contemporaneous and forward (early-access) news,
    \item $\mathcal{P}_{t-1}^{\mathrm{inst}}$ denotes filtered prior-day market commentary,
    \item $\mathcal{S}_{t}^{\mathrm{ret}}$ denotes retail flow statistics (aggregate-level signals),
    \item $\mathcal{F}_{j,t}$ denotes firm-level fundamental information,
    \item $\mathcal{M}_{j,t}$ denotes market state variables (price, volume, order book).
\end{itemize}

\subsection{Hierarchical Multi-Agent Signal Generation}

The institution agent consists of three sub-agents and one aggregator acting as risk manager. Each sub-agent produces a scalar signal in $[-1,1]$ for each asset.

\paragraph{Fundamental Signal}

The fundamental sub-agent evaluates expectation changes in intrinsic value:
\begin{equation}
s^{\mathrm{fund}}_{j,t} = f_{\mathrm{fund}}\left(\mathcal{F}_{j,t}, \mathcal{N}_{t}, \mathcal{N}_{t+1}\right),
\end{equation}
where $s^{\mathrm{fund}}_{j,t} \in [-1,1]$ captures the direction and magnitude of expected valuation revision.

\paragraph{Quantitative Signal}

The quantitative sub-agent extracts signals from technical and capital flow features:
\begin{equation}
s^{\mathrm{quant}}_{j,t} = f_{\mathrm{quant}}\left(\mathcal{M}_{j,t}\right),
\end{equation}
where $s^{\mathrm{quant}}_{j,t} \in [-1,1]$ reflects short-term price dynamics and liquidity conditions.

\paragraph{Sentiment and Flow Signal}

The sentiment/flow sub-agent evaluates retail behavior and crowd positioning:
\begin{equation}
s^{\mathrm{sent}}_{j,t} = f_{\mathrm{sent}}\left(\mathcal{S}_{t}^{\mathrm{ret}}, \mathcal{P}_{t-1}^{\mathrm{inst}}, \mathcal{N}_{t}\right),
\end{equation}
where $s^{\mathrm{sent}}_{j,t} \in [-1,1]$ captures contrarian opportunities and behavioral extremes.

\subsection{Risk Management and Decision}

The risk management agent aggregates sub-agent signals into a unified decision signal:
\begin{equation}
s^{\mathrm{rm}}_{j,t} = g\left(s^{\mathrm{fund}}_{j,t}, s^{\mathrm{quant}}_{j,t}, s^{\mathrm{sent}}_{j,t}, \mathbf{w}_{t}\right),
\end{equation}
where $\mathbf{w}_{t}$ denotes the current portfolio weights.

The risk management agent outputs a target portfolio adjustment:
\begin{equation}
\Delta w_{j,t} = h\left(s^{\mathrm{rm}}_{j,t}, \kappa_t\right),
\end{equation}
where $\kappa_t \in [0,1]$ denotes decision conviction, and $\Delta w_{j,t} \in \mathbb{R}$ is the desired change in portfolio weight for asset $j$.

\subsection{Portfolio Constraints}

The institution agent operates under portfolio-level constraints. Let $w_{j,t}$ denote the portfolio weight of asset $j$ at time $t$. The updated weights satisfy:
\begin{equation}
w_{j,t+1} = \mathrm{clip}\left(w_{j,t} + \Delta w_{j,t}, \, 0, \, w_{\max}\right),
\end{equation}
subject to:
\begin{equation}
\sum_{j=1}^{M} w_{j,t+1} \leq W_{\max}, \quad \sum_{j=1}^{M} w_{j,t+1} \geq W_{\min},
\end{equation}
where $w_{\max}$ is the maximum single-asset weight, and $W_{\max}, W_{\min}$ are total portfolio bounds.

To account for market conditions, adjustments are scaled by expected volatility $\bar{v}_t$:
\begin{equation}
\Delta w_{j,t}^{\mathrm{adj}} = \Delta w_{j,t} \cdot \left(1 - \beta \cdot \max(0, \bar{v}_t - v_0)\right),
\end{equation}
where $\beta > 0$ is a scaling parameter and $v_0$ is a volatility threshold.

\subsection{Order Mapping and Execution}

Let $W_t^{\mathrm{inst}}$ denote total portfolio value and $P_{j,t}$ the asset price. The target number of shares is:
\begin{equation}
Q_{j,t} = \left\lfloor \frac{\Delta w_{j,t}^{\mathrm{adj}} \cdot W_t^{\mathrm{inst}}}{P_{j,t}} \right\rfloor.
\end{equation}

Orders are submitted to the limit order book (LOB). Execution strategy depends on urgency $u_{j,t} \in [0,1]$ and order size relative to market depth:
\begin{equation}
\pi_{j,t} =
\begin{cases}
\text{aggressive}, & u_{j,t} > \tau_u, \\
\text{passive}, &
\begin{aligned}
&u_{j,t} \leq \tau_u \ \text{and}\\
&|Q_{j,t}| \leq \theta D_{j,t},
\end{aligned}\\
\text{TWAP}, & \text{otherwise},
\end{cases}
\end{equation}
where $\tau_u$ is an urgency threshold, $D_{j,t}$ denotes LOB depth, and $\theta$ is a size ratio parameter.

\subsection{Dynamics}

The institution agent repeats this pipeline at each time step, with realized trades impacting both the market state $\mathcal{M}_{j,t+1}$ and retail information signals $\mathcal{S}_{t+1}^{\mathrm{ret}}$, thereby closing the feedback loop between informed trading and market dynamics.

% ==============================================================================
% Part 5: originally content/appendix.tex
% ==============================================================================
\section{Semantic Indistinguishability: Experimental Protocol and Results}
\label{app:semantic_exp}

This appendix provides the full experimental design and results underlying the Semantic Indistinguishability findings summarized in Section~\ref{sec:pilot_controllability}.

\subsection{Experimental Design}
\label{app:exp_design}

Following the behavioral testing methodology of \cite{ribeiro2020beyond}, we design a minimal, single-step, single-asset position-adjustment task that isolates the control interface from confounding factors (price feedback, wealth path dependence, agent interaction).  Each agent receives a fixed market state and selects an action from the discrete action space
\[
\begin{aligned}
\mathcal{A}=\{&-100,-75,-50,-25,0,\\
              &25,50,75,100\},
\end{aligned}
\]
where values represent target position change in percentage points.

\paragraph{Test types.}
\begin{enumerate}
    \item \textbf{MFT (Minimum Functionality Test):} Verifies basic directional sanity---agents should buy in strongly bullish conditions and sell in strongly bearish conditions.
    \item \textbf{DIR (Directional Expectation Test):} Verifies whether an intended risk gradient $L_0<L_1<\cdots<L_9$ induces a monotonic behavioral gradient.
    \item \textbf{INV (Invariance Test):} Verifies whether semantically equivalent paraphrases of the same intended risk level preserve behavioral distributions.
\end{enumerate}

\paragraph{Control interfaces tested.}
\begin{itemize}
    \item \textbf{NL Profile} ($\Phi^{\mathrm{NL}}$): $10$ risk levels $\times$ $10$ paraphrases $=100$ agents.  Each agent's system prompt contains a natural-language personality trait ranging from ``extremely cautious'' ($L_0$) to ``extremely aggressive'' ($L_9$).
    \item \textbf{Numeric Prompt} ($\Phi^{\mathrm{Num}}$): 10 agents with an explicit \\
    \texttt{risk\_tolerance} $\in\{0.0,0.1,\ldots,0.9\}$ in the prompt.
    \item \textbf{Utility Control} ($\Phi^{\mathrm{Util}}$): 10 risk levels mapped to a parameterized mean--variance utility with no LLM decision call.  For risk level $l\in\{0,\ldots,9\}$:
\end{itemize}
\begin{equation}
\label{eq:util_params}
\alpha_l = 0.01+0.01\,l, \qquad \gamma_l = 5.0-0.4\,l,
\end{equation}
\begin{equation}
\label{eq:util_func}
\hat{R} = \alpha_l\cdot s_{\mathrm{composite}},\qquad
U(w) = w\,\hat{R} - \tfrac{\gamma_l}{2}\,w^2\,\hat{\sigma}^2,
\end{equation}
\begin{equation}
\label{eq:util_action}
a^* = 100\cdot\arg\max_{w\in\mathcal{W}} U(w),
\end{equation}
where $s_{\mathrm{composite}}$ and $\hat{\sigma}^2$ are computed from the scenario's z-score signals, and $\mathcal{W}=\{-1.0,-0.75,\ldots,1.0\}$.

All conditions share one fixed minimal-control scenario, constructed to satisfy four pre-registered criteria: (i)~approximate upside--downside balance, (ii)~no single-signal dominance, (iii)~discriminability under structured control, and (iv)~neutral controllers do not induce extreme actions.

\subsection{Metrics}
\label{app:metrics}

For each agent $i$, the empirical action distribution from $K=50$ repeated samples is
\begin{equation}
\label{eq:action_dist}
p_i(a)=\frac{\mathrm{count}_i(a)}{K},\qquad
\mu_i=\sum_{a\in\mathcal{A}}a\cdot p_i(a).
\end{equation}
For each intended level $l$, the level mean is $M_l=\frac{1}{|\mathcal{S}_l|}\sum_{i\in\mathcal{S}_l}\mu_i$.  Monotonicity is measured by Spearman's $\rho(l,M_l)$ and the adjacent inversion rate:
\begin{equation}
\label{eq:inv_rate}
\mathrm{InvRate}=\frac{\#\{l:M_{l+1}\le M_l\}}{|\mathcal{L}|-1}.
\end{equation}
For invariance testing, the behavioral distance between agents $i,j$ is measured by the Jensen--Shannon divergence:
\begin{equation}
\label{eq:jsd}
d_{\mathrm{beh}}(i,j)=D_{\mathrm{JS}}(p_i\|p_j).
\end{equation}
The noise threshold $\tau_{\mathrm{beh}}$ is calibrated via split-half resampling: for each agent, the $K$ samples are randomly split into two halves 100 times, and $\tau_{\mathrm{beh}}$ is set to the 95th percentile of the resulting JSD distribution.  The within-level violation rate is:
\begin{equation}
\label{eq:viol_rate}
\begin{aligned}
\mathrm{ViolRate}
&=\Pr\!\Bigl(D_{\mathrm{JS}}(p_i,p_j)>\tau_{\mathrm{beh}}\\
&\qquad\Bigm|\,\mathrm{level}(i)=\mathrm{level}(j),\ i\ne j\Bigr).
\end{aligned}
\end{equation}

\subsection{Results}
\label{app:results}

\subsubsection{MFT Results}
\label{app:mft}

Even the minimum functionality test reveals prompt-based fragility.  The most conservative NL agent ($L_0$) produces a mean action of $-5.0$ (sell) under a strongly bullish scenario, while the most aggressive NL agent ($L_9$) produces $+96.25$ (strong buy) under a strongly bearish scenario.  Extreme personality traits in prompts override objective market signals, yielding irrational decisions.

\begin{table}[htbp]
\centering
\small
\setlength{\tabcolsep}{3pt}
\begin{tabular*}{\linewidth}{@{\extracolsep{\fill}}lrrrr@{}}
\toprule
Scenario & \shortstack{Pass\\Count} & Total & \shortstack{Pass\\Rate} & \shortstack{All\\Pass} \\
\midrule
Bullish & 4 & 6 & 66.7\% & No \\
Bearish & 5 & 6 & 83.3\% & No \\
\bottomrule
\end{tabular*}
\caption{MFT pass rates.  NL-controlled agents fail basic directional sanity in both bullish and bearish scenarios.}
\label{tab:mft}
\end{table}

\subsubsection{DIR Results}
\label{app:dir}

Monotonicity improves systematically as the control interface moves from semantic to parametric space (Table~\ref{tab:controllability}).
\begin{table}[htbp]
\centering
\footnotesize
\setlength{\tabcolsep}{3pt}
\begin{tabular*}{\linewidth}{@{\extracolsep{\fill}}lccr@{}}
\toprule
Condition & \shortstack{Spearman\\$\rho$} & \shortstack{Adjacent\\Inversion Rate} & Stability \\
\midrule
NL Profile & 0.867 & 33.3\% & Poor \\
Numeric Prompt & 0.956 & 11.1\% & Moderate \\
Utility Control & 0.919 & \textbf{0.0\%} & \textbf{Excellent} \\
\bottomrule
\end{tabular*}
\captionsetup{font=footnotesize, justification=raggedright, singlelinecheck=false}
\caption{Controllability comparison across three interfaces. Utility control achieves perfect monotonicity (zero inversions) with no LLM decision calls.  Referenced in Section~\ref{sec:pilot_controllability}.}
\label{tab:controllability}
\end{table}

\subsubsection{INV Results}
\label{app:inv}

With $\tau_{\mathrm{beh}}=0.0484$, the result is shown in Table~\ref{tab:inv}.

\begin{table}[!htbp]
\centering
\begin{tabular*}{\linewidth}{@{\extracolsep{\fill}} lr @{}}
\toprule
Metric & Value \\
\midrule
Within-level median JSD & 0.0335 \\
Adjacent-level median JSD & 0.1462 \\
\textbf{Within-level violation rate} & \textbf{44\%} \\
\bottomrule
\end{tabular*}
\caption{INV test results.  Nearly half of semantically equivalent paraphrases produce behavioral shifts exceeding the noise threshold.}
\label{tab:inv}
\end{table}

\subsubsection{Semantic--Behavioral Dissociation}
\label{app:dissociation}

\begin{figure}[!t]
    \centering
    \includegraphics[width=\linewidth]{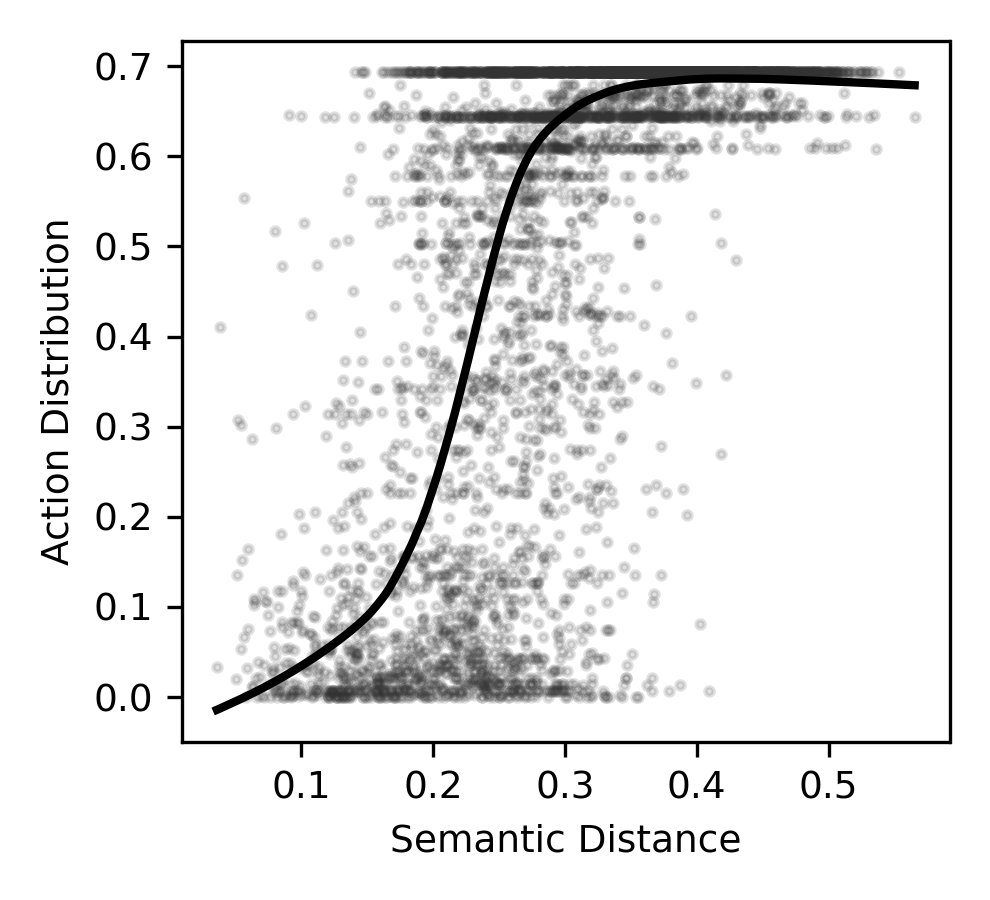}
    \caption{Semantic distance vs.\ behavioral distance for 4{,}950 prompt pairs ($\rho=0.614$).  Referenced in Section~\ref{sec:pilot_controllability}.}
    \Description{Scatter plot comparing semantic distance and behavioral distance across prompt pairs, showing only moderate correlation.}
    \label{fig:sem_beh_scatter}
\end{figure}

Across all 4{,}950 prompt pairs, the Spearman correlation between embedding-space semantic distance and behavioral JSD is only $\rho=0.614$ (Figure~\ref{fig:sem_beh_scatter}).  Semantic proximity does not guarantee behavioral proximity.

\begin{figure*}[t]
    \centering
    \begin{minipage}[t]{0.32\textwidth}
      \centering
      \includegraphics[width=\linewidth]{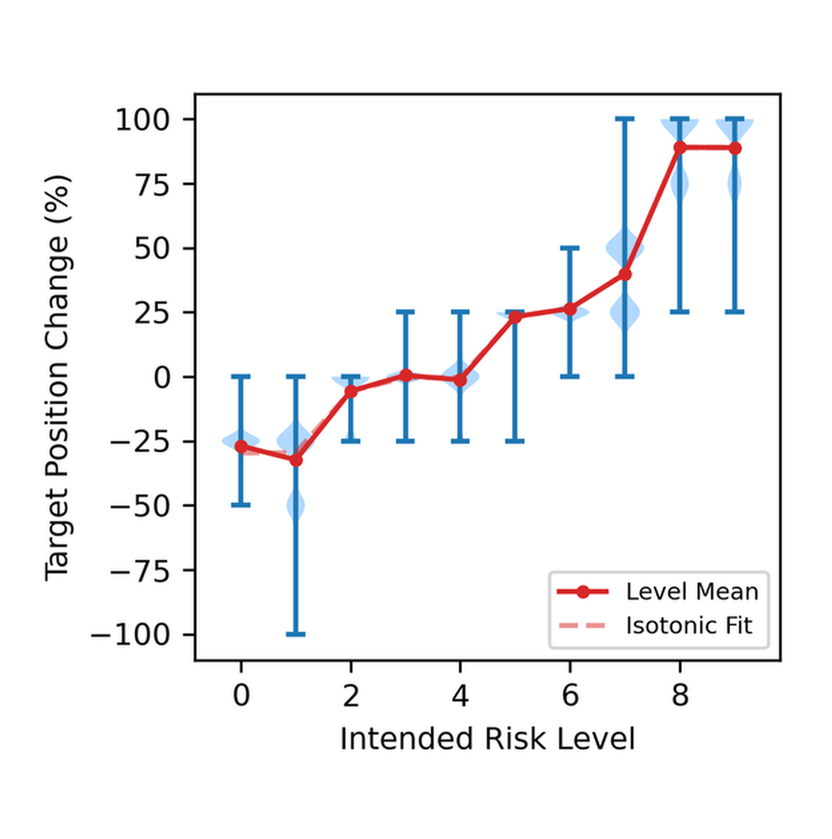}
      \caption*{(a) NL risk-level gradient (L0--L9)}
    \end{minipage}\hfill
    \begin{minipage}[t]{0.32\textwidth}
      \centering
      \includegraphics[width=\linewidth]{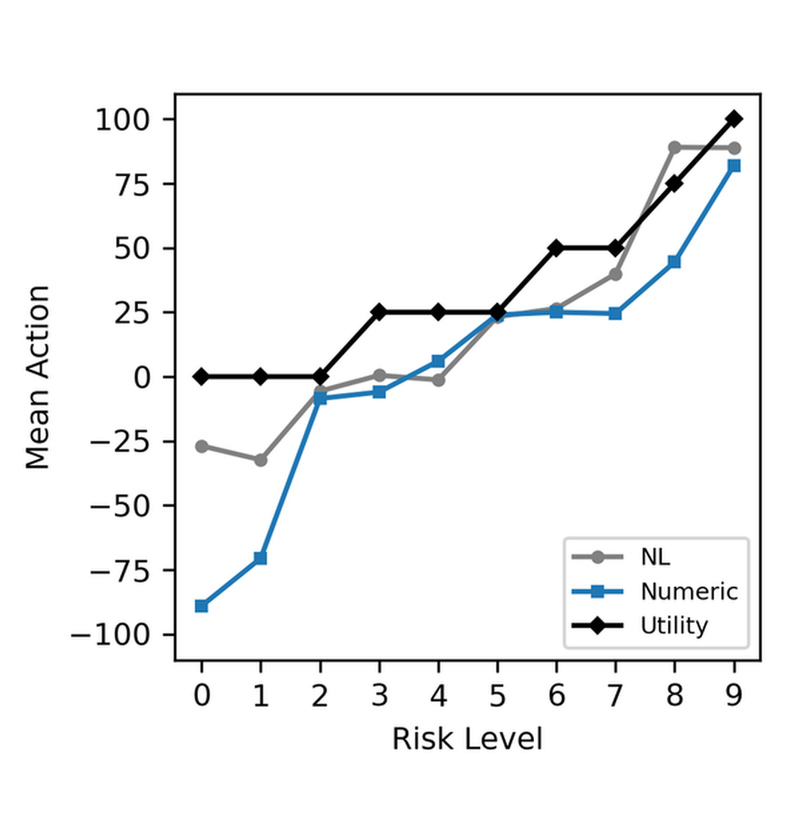}
      \caption*{(b) Three-condition monotonicity}
    \end{minipage}\hfill
    \begin{minipage}[t]{0.32\textwidth}
      \centering
      \includegraphics[width=\linewidth]{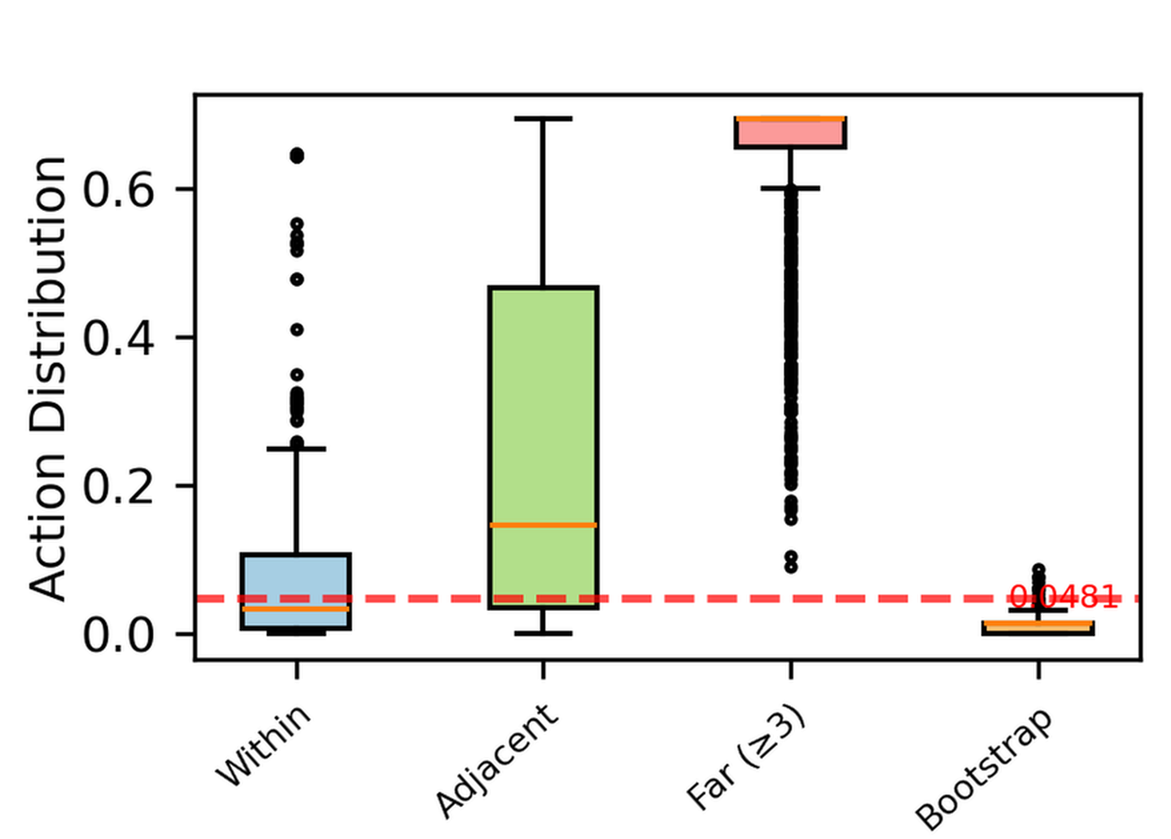}
      \caption*{(c) Within- vs.\ across-level JSD}
    \end{minipage}
    \caption{Semantic indistinguishability diagnostics. \textbf{(a)}~Action distributions across NL risk levels show heavy overlap and frequent local reversals. \textbf{(b)}~Controllability improves from NL to Numeric to Utility; only Utility achieves zero inversions. \textbf{(c)}~Within-level wording variation produces behavioral shifts comparable to genuine risk-level differences (44\% violation rate at $\tau_{\mathrm{beh}}=0.0484$).}
    \Description{Three-panel appendix figure showing overlapping natural-language risk behaviors, improved monotonicity under utility control, and within-level versus across-level Jensen-Shannon divergence.}
    \label{fig:semantic_indist_diagnostic}
\end{figure*}

\section{Forum Mechanism: Endogenous Information Diffusion and Behavioral Amplification}
\label{sec:forum_mechanism}

We introduce a forum mechanism that endogenizes information diffusion among heterogeneous retail agents and provides an observable behavioral signal to institutional agents. The mechanism operates at each discrete time step \(t\) and for each asset \(j\), generating a dynamic text-based information layer that interacts with agents' decision pipelines.

\subsection{Notation and Preliminaries}

Let \(i \in \{1,\dots,N\}\) index agents, \(j \in \{1,\dots,M\}\) index assets, and \(t \in \{1,\dots,T\}\) denote the simulation step. We define:

\begin{itemize}
    \item \(R_{\text{exec},i,j,t} \in \mathbb{R}\): the pre-execution trading signal of agent \(i\), after perception and aggregation but before behavioral adjustment.
    \item \(R_{\text{adj},i,j,t} \in \mathbb{R}\): the post-adjustment signal used for order submission.
    \item \(q_{i,j,t} \in \mathbb{Z}_{\ge 0}\): the inventory (position size) of agent \(i\) in asset \(j\).
    \item \(\rho_{i,t} \in \mathbb{R}\): the latent sentiment state of noise-driven agent \(i\).
    \item \(P\): a forum post object with attributes $$(\text{stock}, \text{day}, \text{intensity}, \text{direction}, \text{author\_type})$$
\end{itemize}

The forum mechanism is parameterized by constants \\
\(\tau_{\text{post}}, D_{\text{window}}, K_{\text{post}}\), and coefficients \(\{\beta_{\cdot}\}\), all fixed across time.

\subsection{Post Generation Mechanism}

\paragraph{Candidate Set.}
At each \((j,t)\), agents eligible to post are selected via:

\begin{equation}
\mathcal{C}_{j,t}
=
\left\{
i \;\middle|\;
\left|R_{\text{exec},i,j,t}\right| > \tau_{\text{post}},\;
q_{i,j,t} > 0
\right\},
\end{equation}

where \(\tau_{\text{post}} > 0\) is a signal threshold. The inventory constraint \(q_{i,j,t} > 0\) ensures that only agents with economic exposure contribute to public discourse, aligning incentives with revealed preferences.

\paragraph{Author Selection.}
From \(\mathcal{C}_{j,t}\), a fixed number of authors are selected based on signal magnitude:

\begin{equation}
\mathcal{A}_{j,t}
=
\operatorname{TopK}\big(\mathcal{C}_{j,t}, K_{\text{author}};\; |R_{\text{exec},i,j,t}|\big).
\end{equation}

Each selected agent \(i \in \mathcal{A}_{j,t}\) produces a single post \(P_{i,j,t}\), where the directional label is:

\begin{equation}
\text{direction}_{i,j,t} = \operatorname{sign}(R_{\text{exec},i,j,t}).
\end{equation}

\paragraph{Endogenous Coupling with Decision Pipeline.}
The posting probability is implicitly governed by the agent-specific signal formation process:

\begin{align}
R_{1,i,j,t} &= (1+\eta_{i,t}) \hat{R}_{i,j,t}, \\
R_{2,i,j,t} &= (1-h_{i,t}) R_{1,i,j,t} + h_{i,t} r_{\text{consensus},j,t}, \\
R_{\text{exec},i,j,t} &= R_{2,i,j,t} + \epsilon_{i}, \quad \epsilon_i \sim \mathcal{N}(0,\sigma_i^2),
\end{align}

where \(\eta_{i,t}\) captures overreaction, \(h_{i,t}\) captures herding intensity, and \(\sigma_i\) controls idiosyncratic noise. Thus, forum participation is an emergent outcome of heterogeneous behavioral parameters.

\subsection{Post Consumption and Recommendation}

\paragraph{Candidate Pool.}
Agents consume posts from a rolling window:

\begin{equation}
\mathcal{P}_j^{(t)}=
\left\{P\;\middle|\;
\begin{gathered}
P.\text{stock}=j,\\
t-D_{\text{window}}\le P.\text{day}<t
\end{gathered}
\right\}.
\end{equation}

Each post is weighted by a recency decay:

\begin{equation}
\operatorname{recency}(P)
=
0.5^{\max(t - P.\text{day} - 1,\,0)}.
\end{equation}

\paragraph{Reader-Specific Scoring Functions.}
Each agent ranks posts via a type-dependent scoring function:

\begin{equation}
s_R(P,i) = f_R(P,i)\cdot \operatorname{recency}(P),
\end{equation}

where \(R \in \{\text{noise}, \text{momentum}, \text{value}\}\).

\textbf{Noise-driven agents (sentiment reinforcement):}
\begin{equation}
\begin{aligned}
&f_{\text{noise}}(P,i)\\
&\quad=P.\text{intensity}\Bigl(1+\beta_{\text{echo}}\cdot\\
&\qquad\mathbf{1}[\mathrm{dir}(P)=\operatorname{sign}(\rho_{i,t-1})]\Bigr).
\end{aligned}
\end{equation}

\textbf{Momentum agents (technical amplification):}
\begin{equation}
\begin{aligned}
&f_{\text{mom}}(P,i)=P.\text{intensity}\\
&\quad\times\Bigl(1+\beta_{\text{tech}}\cdot
\mathbf{1}[\mathrm{type}(P)=\mathrm{mom}]\Bigr).
\end{aligned}
\end{equation}

\textbf{Value agents (fundamental and contrarian filtering):}
\begin{equation}
\begin{aligned}
&f_{\text{val}}(P,i)=
\beta_{\text{fund}}\mathbf{1}[P.\text{author\_type}=\text{value}]\\
&\quad+\beta_{\text{contra}}\mathbf{1}[P.\text{direction}\ne\operatorname{sign}(R^{\text{prev}}_{i,j})]\\
&\quad+\beta_{\text{base}}P.\text{intensity}.
\end{aligned}
\end{equation}

where \(R^{\text{prev}}_{i,j}\) denotes the agent's prior-period directional signal.

\paragraph{Top-K Injection.}
Each agent receives a truncated feed:

\begin{equation}
\text{Posts}_{i,j,t}
=
\operatorname{TopK}\left(
\mathcal{P}_{j}^{(t)},\;
K_{\text{post}},\;
s_R(P,i)
\right),
\end{equation}

which is injected into the agent's LLM prompt alongside exogenous information (e.g., news).

\subsection{Institutional Interpretation Layer}

Institutional agents do not observe latent retail states directly but access an aggregated representation of forum activity. Specifically, institutions reuse the value-based ranking:

\begin{equation}
\text{Posts}^{\text{inst}}_{j,t}
=
\operatorname{TopK}\left(
\mathcal{P}_{j}^{(t)},\;
K_{\text{post}},\;
s_{\text{val}}(P,i)
\right).
\end{equation}

Additionally, institutions observe summary statistics:

\begin{align}
n_{\text{hold},j,t} &:\text{ number of retail holders}, \\
\Delta n_{j,t} &= n_{\text{buy}} - n_{\text{sell}}, \\
\text{cost\_vs\_price}_{j,t} &= \frac{\overline{\text{cost}}_{j,t}}{P_{j,t}} - 1, \\
(n_{\text{bull}}, n_{\text{bear}}, n_{\text{neutral}}) &: \text{post sentiment counts}.
\end{align}

These features are processed by a sentiment-analysis sub-agent to produce structured signals:

\begin{equation}
\left(
\text{extreme}_{j,t},
\text{pain}_{j,t},
\text{contra}_{j,t}
\right)
\in \mathbb{R}^3
\end{equation}
, which quantify sentiment extremity, unrealized loss pressure, and contrarian opportunity, respectively. These signals are subsequently integrated into the institutional decision hierarchy.

\subsection{Forum Dynamics}

The forum mechanism introduces a bidirectional coupling between decision-making and information generation. On the production side, posting behavior reflects endogenous belief strength and inventory exposure. On the consumption side, heterogeneous filtering functions induce selective attention and echo dynamics. This dual structure enables the emergence of behavioral phenomena such as sentiment cascades, herding amplification, and contrarian opportunities, while preserving a computational complexity that scales linearly with the number of agents.
% ============================================================
% Appendix C--E: Retail Investor Type Specifications
% Insert after existing Appendix B in main document
% ============================================================

\section{Scalability Projection Method}
\label{app:projection_method}

The measured call counts for SA and TM grow approximately in proportion to
$N$ over the three evaluated scales ($N\in\{50,100,500\}$), as shown in
Table~\ref{tab:projection_linearity}. Figure~\ref{fig:resource_cost} reports
the measured token totals. Resource estimates at $N=1000$ assume linear
scaling and double the corresponding $N=500$ value:

\begin{equation}
\label{eq:projection}
\begin{aligned}
C(N=1000)
&=C(N=500)\times\frac{1000}{500}\\
&=2\times C(N=500),
\end{aligned}
\end{equation}

and equivalently for token counts and wall-clock time.
These are projections, not measured results. Actual token use and runtime
can vary with generated outputs and execution settings.

\begin{table}[H]
  \centering
  \scriptsize
  \setlength{\tabcolsep}{3pt}
  \renewcommand{\arraystretch}{1.04}
  \begin{tabular}{@{}lrrr@{}}
    \toprule
    $N$ & \textbf{SA calls} & \textbf{ASD} & \textbf{Calls/ASD} \\
    \midrule
    50  & 1{,}074  & 1{,}500  & 0.716 \\
    100 & 2{,}166  & 3{,}000  & 0.722 \\
    500 & 10{,}710 & 15{,}000 & 0.714 \\
    \bottomrule
  \end{tabular}
  \captionsetup{font=footnotesize, skip=4pt}
  \caption{Measured SA call-count linearity.
  The near-constant Calls/ASD ratio (0.714--0.722) supports
  approximately linear growth over the measured scales.
  The $N=1000$ estimate remains an extrapolation.
  TM figures are identical.}
  \label{tab:projection_linearity}
\end{table}

% ---
\section{Time Period Sensitivity Analysis}
\label{app:time_period_sensitivity}

To assess robustness beyond the time window reported in the main text, we repeat the index-level comparison on the second half of 2025. We construct an equal-weight composite index from the simulated closing prices of the 50-stock universe and compare it against the equal-weight index computed from real market closes over the same trading days, using the same normalization and evaluation protocol as in the first-half experiment.

The second-half results show that the simulator preserves the correct low-frequency direction and magnitude while remaining stable under the shift in time period. Over 126 trading days, the simulated composite index delivers an 8.35 percent return versus 6.70 percent in the real market, indicating that the model captures a comparable medium-horizon drift. At the same time, the path error is moderate, with an RMSE of 17.39 on a 1000-point normalized index, suggesting that the overall trajectory is not dominated by accumulated bias.

However, the day-to-day comovement remains weak: the correlation between simulated and real daily returns is approximately zero. This indicates that, while the simulator can match coarse-grained index dynamics over the horizon, it does not consistently reproduce the exact timing of short-horizon shocks and reversals. Taken together, these findings support the claim that the framework is not overfit to a single evaluation window and generalizes at the index level across periods, while also highlighting a clear limitation in high-frequency alignment that motivates future improvements in shock transmission and calibration.

% ==============================================================================
% Part 6: originally ethics_checklist.tex
% ==============================================================================
\section{Ethical Considerations}

This work studies LLM-assisted financial-market simulation. The proposed system is intended only for academic research on controllable and scalable market simulators. It is not designed, validated, or released for live trading, investment recommendation, portfolio allocation, market manipulation, or any form of automated financial decision-making. All empirical claims should therefore be interpreted as evidence about simulation behavior under a controlled experimental setting, rather than as claims about real-world trading profitability or deployable investment performance.

\paragraph{Potential misuse.}
A market simulator can be misused if readers overinterpret simulated trajectories as trading signals, use simulated agent behavior to justify investment strategies, or treat synthetic market dynamics as a substitute for regulated financial advice. This risk is especially relevant because the simulator combines financial data, textual market information, and LLM-generated semantic interpretation. To reduce this risk, we explicitly restrict the intended use of SUM to research analysis, simulator evaluation, and methodological comparison. We do not present any trading strategy, trading recommendation, expected return forecast, or investment advice. We also report fidelity limitations and avoid claiming that the simulator fully reproduces order-flow realism or real-market causality.

\paragraph{Market-impact and dual-use considerations.}
The framework could in principle be extended by third parties toward strategic trading simulation, stress testing of market narratives, or behavioral manipulation studies. Such use would require additional safeguards, market-impact review, and compliance checks. Our experiments are conducted in an offline historical simulation environment and do not interact with live exchanges, brokers, order-routing systems, or real investors. The simulated orders are synthetic and cannot affect real market prices or liquidity.

\paragraph{Data and privacy.}
The study uses financial-market data, public news or textual market information, LLM API outputs, and simulation logs. We do not intentionally collect private personal data, user-level trading records, brokerage account information, or data from human participants. Public financial news and forum-like textual inputs may still contain names of public figures, company officers, analysts, or quoted individuals. We therefore treat all raw text as potentially sensitive and use it only for research-oriented semantic processing. Where applicable, preprocessing removes irrelevant, duplicate, private, or offensive content before simulation. The released artifacts should avoid redistributing raw third-party text unless redistribution is permitted by the corresponding data license or terms of use; derived features, aggregate statistics, prompts, and code are safer release targets.

\paragraph{Licensing and terms of use.}
The artifacts used in this work include LLM APIs, financial-market data sources, public news or textual information, and simulation code. All such artifacts should be used consistently with their original access conditions. If any commercial data vendor, licensed financial database, or API-restricted source is used, the raw data should not be redistributed unless the license explicitly permits redistribution. When the code is released, the repository should document the data sources, required access credentials, package versions, model/API settings, and any restrictions on reproducing the full experiments. Any public release should separate redistributable code and configuration files from non-redistributable raw market or news data.

\paragraph{Human subjects.}
This work does not involve human-subject experiments, crowdworkers, annotators, surveys, interviews, or intervention on real users. No human participants are recruited, paid, profiled, or evaluated. Therefore, human-subject consent, participant compensation, demographic reporting, and institutional review procedures for human-subject research are not applicable.

\paragraph{Use of AI assistants.}
AI assistants were used for writing refinement, formatting assistance, and language polishing. They were not used as authors. The scientific framing, experimental design, implementation, result interpretation, and final conclusions were checked and approved by the human authors. Any AI-assisted text was reviewed by the authors before submission.

\end{document}